\documentclass[preprint,12pt]{elsarticle}

\usepackage{amssymb}
\usepackage{amsmath}
\usepackage{color}
\usepackage{framed}
\definecolor{shadecolor}{rgb}{0,1,0}
\usepackage{multirow}
\usepackage{subfigure}

\journal{Pattern Recognition}

\begin{document}
	
	\begin{frontmatter}
		
		
		
		\title{Joint-Individual Fusion Structure with Fusion Attention Module for Multi-Modal Skin Cancer Classification}

		\author[label1,label2]{Peng TANG}
		\author[label3]{Xintong YAN}
		\author[label4]{Yang NAN}
		\author[label1,label5]{Xiaobin HU}
		\author[label1,label5]{Xiaobin HU}
		\author[label5]{Bjoern H. Menze}
		\author[label6]{Sebastian Krammer}
		\author[label1,label2]{Tobias Lasser}

		\address[label1]{organization={Department of Informatics, School of Computation, Information, and Technology, Technical University of Munich},
			city={Garching},
			country={Germany}}
		
		\address[label2]{organization={Munich Institute of Biomedical Engineering, Technical University of Munich},
			city={Garching},
			country={Germany}
		}
		
		\address[label3]{organization={Henan Economic Research Institute, State Grid Corporation of China},
			city={Zhengzhou},
			country={China}}
		
		\address[label4]{organization={National Heart and Lung Institute, Imperial College London},
			city={London},
			country={UK}}
		
		\address[label5]{organization={Department of Quantitative Biomedicine, University of Zurich},
			city={Zurich},
			country={Switzerland}}
		
		\address[label6]{organization={Department of Dermatology and Allergy, University Hospital, LMU Munich},
			city={Munich},
			country={Germany}}
		
		\begin{abstract}

			Most convolutional neural network (CNN) based methods for skin cancer classification obtain their results using only dermatological images.  
			Although good classification results have been shown, more accurate results can be achieved by considering the patient's metadata, which is valuable clinical information for dermatologists.
			Current methods only use the simple joint fusion structure (FS) and fusion modules (FMs) for the multi-modal classification methods, there still is room to increase the accuracy by exploring more advanced FS and FM.  
			Therefore, in this paper, we design a new fusion method that combines dermatological images (dermoscopy images or clinical images) and patient metadata for skin cancer classification from the perspectives of FS and FM.
			First, we propose a joint-individual fusion (JIF) structure that learns the shared features of multi-modality data and preserves specific features simultaneously.
			Second, we introduce a fusion attention (FA) module that enhances the most relevant image and metadata features based on both the self and mutual attention mechanism to support the decision-making pipeline.
			We compare the proposed JIF-MMFA method with other state-of-the-art fusion methods on three different public datasets. 
			The results show that our JIF-MMFA method improves the classification results for all tested CNN backbones and performs better than the other fusion methods on the three public datasets, demonstrating our method's effectiveness and robustness

		\end{abstract}
		
		
%
		
	\end{frontmatter}
	
	
	
	\section{Introduction}
	Skin cancer is one of the most dangerous and fast-growing cancers in the world \citep{WHO2020}.
	In the US, the annual number of skin cancer's estimated new cases exceeded 110,000 and estimated death cases is over 11,000 \citep{ CS2021} during the past five years.
	During routine examination of skin cancer, typically magnified images of the skin lesions and the patient's meta-data are collected. 
	Afterwards, diagnosis decisions are made based on these two kinds of clinical information and the doctor’s experience \citep{pacheco2020impact}.   
	On the one hand, accurate diagnosis is challenging. 
	It relies on appropriate training and experience in the dermoscopic images that are obtained by a non-invasive imaging technique. 
	This technique enlarges and illuminates skin lesions to show visual features of deep skin that are not visible to the human eye \citep{pacheco2021attention, argenziano2003dermoscopy}.
	On the other hand, even for the experienced dermatologists, the diagnosis is potentially affected by stress, fatigue, or other human factors, and thus the same high diagnosis accuracy is not guaranteed every single time. 
	Therefore, a computer-aided diagnosis system for skin cancer is expected to aid even very experienced dermatologists and to help improve the overall diagnostic outcome \citep{pacheco2021attention}.
	
	\subsection{Skin cancer classification based on dermatological images} 
	Recently, deep learning based models have dominated the field of medical image analysis, including skin lesion classification.
	This is evidenced by the fact that convolutional neural networks (CNNs) and their variants have won almost all the skin lesion classification (SLC) challenges hosted by International Skin Imaging Collaboration (ISIC) since 2016.
	For example, \cite{yu2016automated} introduced a very deep residual CNN for melanoma classification, yielding the best performance in the ISIC 2016 SLC challenge among all the 25 teams. 
	Also, in recent publications, many state-of-the-art methods for the SLC task were developed based on CNN \citep{tang2020GPCNN, tang2019efficient, xie2020mutual,yang2018clinical,jin2021cascade}.
	\cite{tang2020GPCNN} proposed a GP-CNN-DTEL framework, which obtained high scores in the classification of melanoma and nevi, capturing global-local information and adopting data-transformed ensemble learning.
	\cite{xie2020mutual} presented a mutual bootstrap method to boost the performance of both segmentation and classification models, achieving a Jaccard index of 80.4$\%$ in the segmentation task and an average AUC value of 93.8$\%$ in the classification task on the dataset of the ISIC 2017 challenge.
	\cite{wang2021knowledge} designed a cascade knowledge diffusion network that transferred and fused features learned from the model of skin lesion segmentation to get more accurate SLC results.
	\cite{ge2017skin} and \cite{bi2020multi} introduced works that combined both dermoscopy and clinical images for skin disease recognition. 
	In addition, some works \citep{esteva2017dermatologist, haenssle2018man, brinker2019deep,barata2021explainable} conducted performance comparisons between CNN based methods and medical experts. 
	These studies reported that CNNs achieved similar or higher diagnostic accuracy compared to most dermatologists.
	However, these works only used dermoscopy images for the skin disease recognition and did not take the patient metadata into consideration.
	
	\subsection{Skin cancer classification based on the fusion of dermatological images and patient metadata}
	Multi-modal information fusion denotes the task of fusing multiple types of data from different sources \citep{atrey2010multimodal}. 
	It aims to capture supplementary and complete information for better performing machine learning algorithms compared to only using single modality data \citep{huang2020fusion}.
	Over the past years, multi-modal deep learning models have been successfully applied in fields outside of medical image analysis \citep{huang2020fusion}.
	For example, \cite{trzcinski2018multimodal} adopted a multi-modal pipeline that combines visual and textual features for social media video classification.  
	The classification accuracy of 76.4$\%$ obtained by single modality CNN was increased to 88$\%$ by this multi-modal pipeline.
	\cite{person2019multimodal} designed a detection system that fuses images with other data from Light Detection and Ranging (LiDAR) sensors for autonomous driving. 
	This detection system achieves 3.7$\%$ higher accuracy than the detection model trained by single modality data.
	These successful applications of multi-modal information fusion attracted the attention of many researchers who are working with medical applications.
	Leveraging a multi-modal fusion scheme is also expected to provide complementary information and overcome the limitations of single-modality models. 
	According to a recent literature review \citep{huang2020fusion}, there is a trend that the integration of image and electronic health records is used to solve many tasks that can not be robustly handled by single-modality models in the field of medical image analysis, including dermatological image analysis.
	
	Patient's demographics are important clinical information during dermatologists' examinations.
	Especially in cases where visual features of skin lesions have inter-class similarity and intra-class difference, clinical metadata (such as age, gender, lesion's location, parents' background, skin cancer history and others) became crucial factors for dermatologists diagnosis \citep{pacheco2021attention}.
	Many studies combining dermoscopy images and patient metadata have been reported \cite{yap2018multimodal, kawahara2019seven, liu2020deep, pacheco2021attention, li2020fusing}.
	To the best of our knowledge, \cite{yap2018multimodal} proposed the first work that used a deep learning based model to combine two-modality dermatological images and patient meta-data for skin lesion classification.
	\cite{kawahara2019seven} introduced a multi-modal learning network that fuses images data and patient metadata for multi-label skin lesion classification. 
	\cite{liu2020deep} designed a deep learning based system that combined multi-view images and metadata for differentiating skin diseases. 
	The deep learning system achieved comparable performance with dermatologists but outperformed six primary care physicians and six nurse practitioners in their validation.
	Although these works presented notable results, these methods only integrated the data of two modalities by feature concatenation, which may not capture the latent relationship between dermatological images and metadata \citep{pacheco2020impact, li2020fusing}.
	More recently, some researchers believed that the concatenation operation could not make full use of multi-modal data. 
	Some works \cite{pacheco2021attention,liu2020deep, cai2022multimodal}, i.e. Metablock, Metanet and Mutual Attention, Met proposed to extract the most relevant image features by an attention-based mechanism, which achieved better performance than the concatenation operation.
	
	However, these approaches mentioned above generally used the joint fusion structure to fuse images and patient metadata. 
	This means that these methods only learn a joint feature representation of multi-modality data and neglect to retain the specific characteristics of each modality that has been verified to be crucial for the multi-modal task \citep{he2021multi, hu2017sharable}.
	Also, most of current fusion modules (fusion operations) only used the metadata to enhance the most relevant image features, and did not explore the possibility of using both image and metadata to enhance the most-related features of both these two-modality data. 
	Therefore, in our opinion, there still exists great potential to get more accurate results by designing an improved fusion approach regarding the overall structure and a multi-modal attention module.
	
	In this work, we introduce a joint-individual fusion (JIF) structure with a multi-modal mutual (MMFA) attention module to integrate the dermatological images and patient metadata.
	First, the JIF structure jointly learns an improved shared multi-modal feature representation through preserving modal-specific features.
	Second, the MMFA attention module is designed to enhance the most relevant image features and metadata features, where the most relevant features of a modality will be highlighted by the other modality and its own features.   
	The proposed method was evaluated on three public datasets, the \textit{i.e.} PAD-UFES-20 \citep{PACHECO2020106221}, Seven-Point Check like (SPC) \citep{kawahara2019seven} and ISIC-2019 datasets \citep{tschandl2018ham10000, codella2018skin, combalia2019bcn20000}, and compared with other currently state-of-the-art fusion methods: Joint Fusion (JF) structure with Concatenation, Metanet and Metablocks \citep{pacheco2020impact,li2020fusing, pacheco2021attention}.
	The experimental results demonstrate that our JIF-MMFA method is capable of consistently improving the performance of different CNNs and generally performs better than the other fusion methods on the different datasets.
	
	The method contribution can be summarized as follows:
	1. Compared to previous methods that only focus on a developing new fusion module, we add an idea to improve the performance by exploring more efficient fusion structures.
	2. a new JIF structure that learns modal-shared and modal-specific features simultaneously, which can consistently improve the classification performance of different fusion modules, different backbones and different datasets.
	3. a new FA module that enhances the most relevant image features and metadata features, where the most relevant features of a modality will be highlighted by the the features from the other modality and itself.
	4. Taking advantage of JIF structure and FA module, we contribute a JIF-MMFA method, which achieves state-of-the-art performance on multiple skin diseases datasets.
	
	\begin{figure*}
		\centering
		\includegraphics[height=8cm,width=14cm]{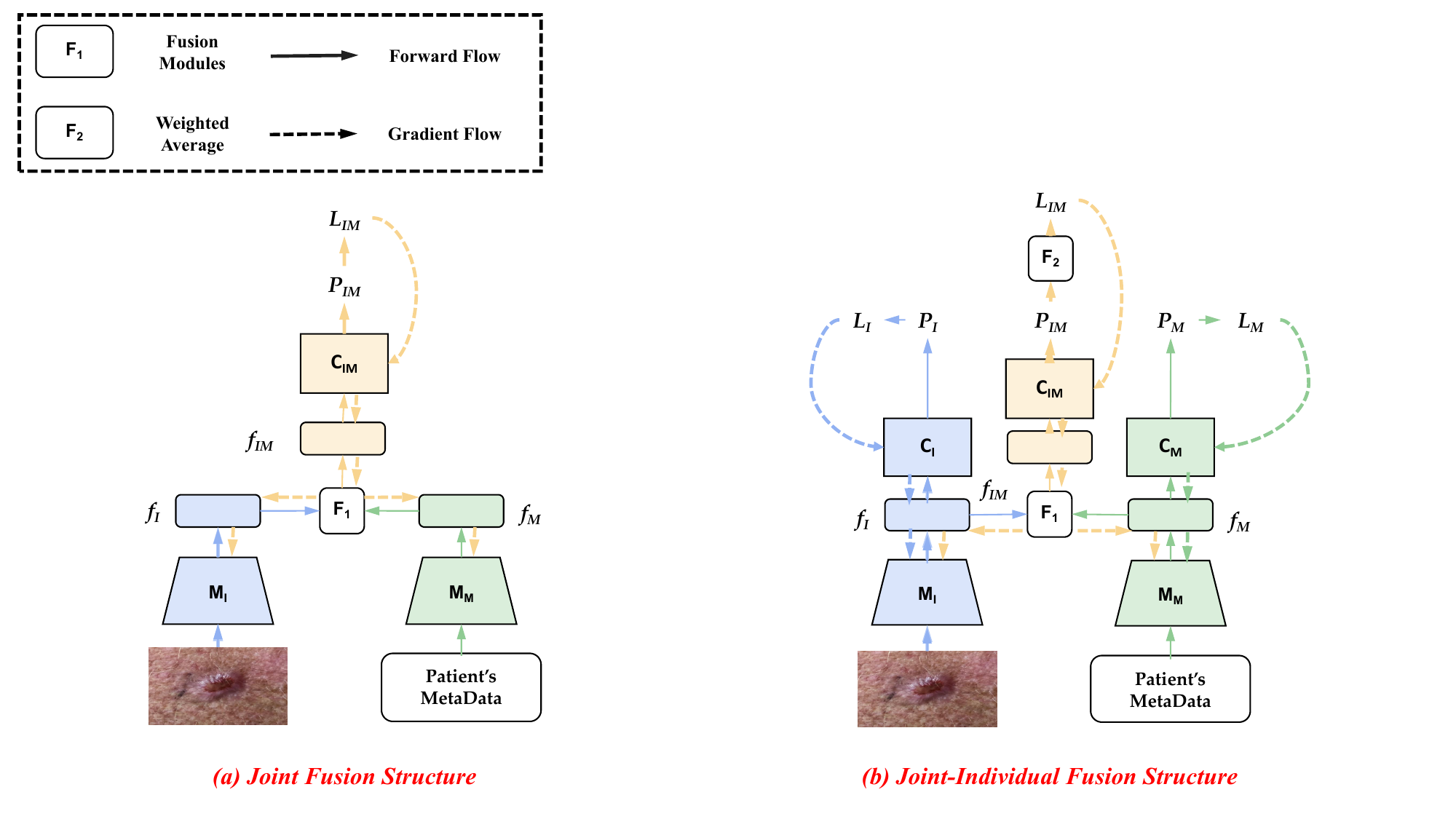}

		\caption{Overview of joint fusion structure (a) and joint-individual fusion structure (b), see also sections 2.1 and 2.2. 
			In this figure, the dermatological image branch is marked in blue, the patient metadata branch is marked in green, and the fusion branch is marked in yellow. 
			The corresponding forward and gradient flows of these three branches are also marked in the corresponding color.  
			$M_I$ is the model to extract image features; $M_M$ is the method to extract patient metadata features; $f_I$, $f_M$ and $f_{IM}$ are the extracted image features, the extracted metadata features, and the features integrated by $f_I$ and $f_M$, respectively. 
			$C_I$, $C_M$, and $C_{IM}$ are the corresponding classifiers of $f_I$, $f_M$, and $F_{IM}$, respectively. 
			$P_I$, $P_M$, and $P_{IM}$ are the predictions obtained from $C_I$, $C_M$, and $C_{IM}$, respectively.
			$L_I$, $L_M$, and $L_{IM}$ are the corresponding loss functions for $C_I$, $C_M$, and $C_{IM}$, respectively.
			In this workflow, the inputs are the dermatological image and the patient metadata, and the output are the predictions $P_I$, $P_M$, and $P_{IM}$.}

		\vspace{-0.2in}
		\label{fig1}
	\end{figure*}
	
	\section{Method}
	\subsection{Notation}
	For convenience, we consider the fusion of dermatological images and patient metadata for skin cancer recognition as a multi-class classification task, in which each case consists of an image $Image$, a group of patient metadata $Meta$, and a ground truth label $GT \in \{1,2,3,..., N\}$, where $N$ is the number of labels. 
	We also build feature extractors $M_I$ and $M_M$ to process the raw data $I$ and $M$. 
	For the raw image data $I$, the CNN $M_I$ is employed to extract the image features  $f_I \in \mathbb{R}^{D_I}$ (the last-layer feature maps of the CNN), where $D_I$ is the dimension of the image features $f_I$. 
	Regarding the patient metadata, one-hot encoding and multiple fully connected layers (FCLs) are adopted as $M_M$ to transform the raw data into non-linear metadata features $f_M\in \mathbb{R}^{D_M}$, where $D_M$ is the dimension of the metadata features $f_M$.
	These two feature extraction processes can be formulated as: 
	\begin{equation}
		f_I = M_{I}(Image)
	\end{equation}
	\begin{equation}
		f_M = M_{M}(Meta)
	\end{equation}
	
	Thus, our goal is to propose a method ($\text{ME}$) that predicts the probability $P$ of $GT$ assuming a class $c \in \{1,2,3,...,N\}$ given the image $I$ and metadata $M$:
	\begin{equation}
		P_{GT} = \text{ME}(~GT~=~c~|~f_I,~f_M)
		\label{eq3}
	\end{equation}

	\subsection{Joint-Individual Fusion (JIF) structure}
	To describe the former methods based on the Joint Fusion structure (see Fig.~\ref{fig1}), Eq.~(\ref{eq3}) is modified as follows:

	\begin{equation}
		 P_{IM} := C_{IM}(~GT~=~c~|~F_1(~f_I,~f_M)~)
		\label{eq4}
	\end{equation}
	
	where $C_{IM}$ is the corresponding classifier of the fused features $F_{IM}$, where $F_{IM} = F_1(~f_I,~f_M)$, and $F_1$ indicates the fusion module, and $P_{IM}$ is the prediction from $C_{IM}$.
	
	For the proposed Joint-Individual Fusion structure, Eq.~(\ref{eq3}) is further derived as follows:
	
	\begin{equation*}
		P_{IM} := C_{IM}(~GT~=~c~|~F_1(~f_I,~f_M)~) 
	\end{equation*}
	\begin{equation*}
		P_{I} := C_{I}(~GT~=~c~|~f_I~) 
	\end{equation*}
	\begin{equation*}
		P_{M} := C_{M}(~GT~=~c~|~f_M~)
	\end{equation*}
	\begin{equation}
		P_{GT} := F_2(~P_{IM},~P_{I},~P_{M})
		\label{eq5}
	\end{equation}
	
	where $C_I$ and $C_M$ are the corresponding classifiers of the image features $f_I$ and metadata features $f_M$. 
	$P_{I}$ and $P_M$ are the predictions of $C_{I}$ and $C_{M}$. 
	$C_{IM}$ is a fully connected layer that is commonly used as classifier by CNNs to predict the last feature maps.
	From Eq.~(\ref{eq4}) and Eq.~(\ref{eq5}),  it can be seen that the main differences between the proposed JIF structure and the JF structure are in $F_2$, $P_I$, and $P_M$. 
	These differences can be further differentiated by two aspects: training and testing.
	
	During the training, we adopt an intuitive way where two loss functions $L_I$ and $L_M$ are added to individually supervise the image branch ($C_I$ and $FM_I$) and metadata branch ($C_M$ and $FM_M$).
	The whole gradient flow is changed to enable the model to learn a joint feature representation while retaining the specific features of each modality.
	As visualized in Fig.~\ref{fig1}(b), the gradient from $L_I$ (blue) and $L_M$ (green) guide the image branch and metadata branch to preserve the specific representations $f_M$ and $f_I$, respectively.
	$L_{IM}$ optimizes the whole structure and thus obtains the joint feature representation $F_{IM}$.
	
	During testing, since there are three classifiers in the JIF structure, we naturally integrate these three predictions at the decision level for more accurate results.
	
	\begin{figure}[h]
		\centering
		\includegraphics[height=8cm,width=14cm]{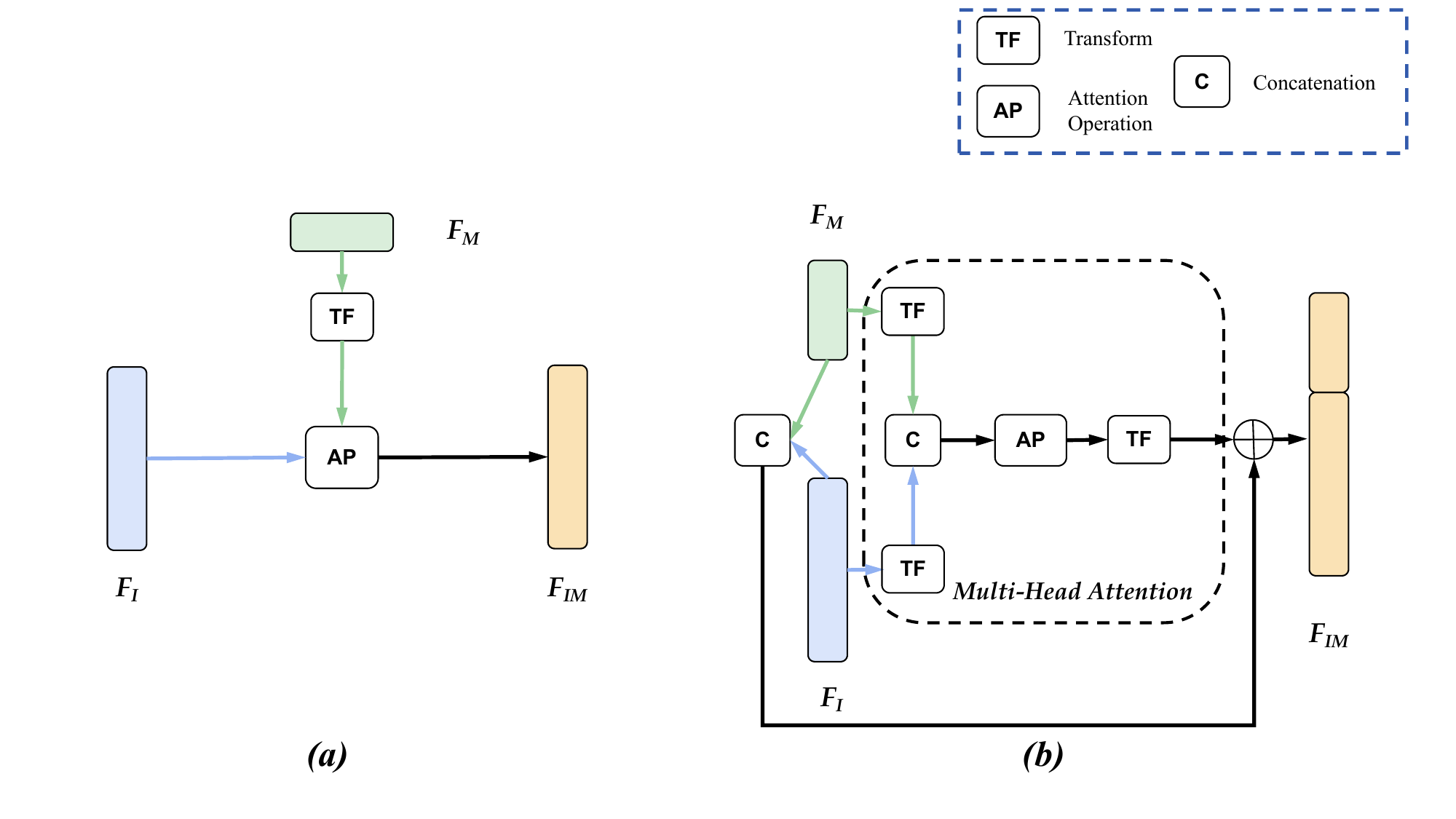}
		\caption{Overview of (a) Metablock and MetaNet, in which the metadata ($f_M$) is used to enhance the image features ($f_I$), and (b) our proposed multi-modal fusion attention module, in which both image features ($f_I$) and metadata features ($f_M$) are enhanced by the features of other modality data and its own features. 
			TF indicates the transformation operation, a single-layer neural network in our experiment. 
			AP refers to the attention operations, such as element-wise multiplication and summation, and self-attention. 
			C is the concatenation operation. 
			$F_{IM}$ is the enhanced feature representation after the fusion module fuses $f_I$ and $f_M$, see also section 2.3.}
		\vspace{-0.2in}
		\label{fig2}
	\end{figure}
	
	\begin{figure}[h]
		\centering
		\includegraphics[height=8cm,width=14cm]{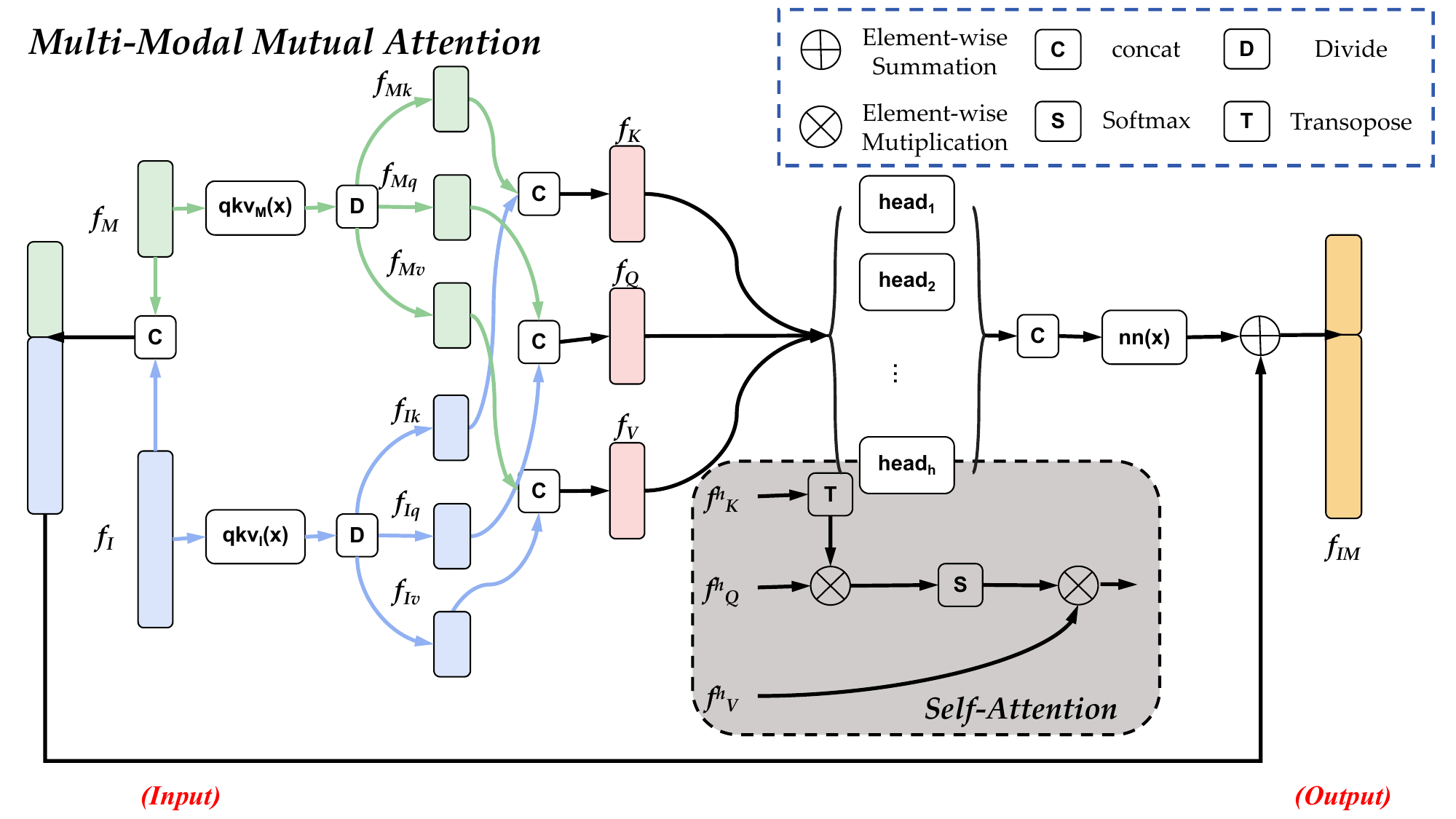}
		
		\caption{
			The structure of the Multi-Modal Fusion Attention (MMFA) module. 
			The MMFA module learns how to enhance both image features ($f_I$) and metadata features ($f_M$) based on their own features and other modality features simultaneously. 
			The length of output feature $f_{IM}$ is the sum of the length of $f_I$ and $f_M$ in our MMFA module.
			$qkv$ is a single-layer neural network,
			$f_{IM}$ is the enhanced feature.
			$f_{Mk}$, $f_{Mq}$, $f_{Mv}$, $f_{Ik}$, $f_{Iq}$, $f_{Iv}$, $f_K$,$ f_Q$, $f_V$, $f^h_K$, $f^h_Q$, $f^h_V$ are the intermediate feature vectors in the attention mechanism, the details about them can be seen in the literature \citep{vaswani2017attention, dosovitskiy2020image}, see also section 2.3.}
		
		\vspace{-0.2in}
		\label{fig3}
	\end{figure}

	\subsection{Multi-Modal Fusion Attention (MMFA) module}

	The proposed fusion module aims to enhance the patient metadata features and the image features by the features of both two-modality data. For example, the image features can be enhanced not only by metadata features but also image features themselves simultaneously, since, from the perspective of data-driven learning, more information can be integrated into the attention/fusion operation, more task-related features can be selected to improve the performance of skin lesion classification.

	The MMFA module implements both mutual attention and self attention mechanism to this two-modality data to get the enhanced fusion feature vectors.
	Then, it uses a skip connection to $f_I$, $f_M$, and the enhanced feature vectors to construct the final fused feature vector $f_{IM}$. 
	This skip connection can avoid the vanishing gradient problem \citep{he2016deep} and can also exploit the useful information in the original feature vectors.
	
	A structure overview of the proposed MMFA module is shown in Fig.~\ref{fig3} and can be summarized as follows:
	\begin{equation}
		F_{IM} := F_1(,f_M)=\text{MMFA}(, f_M) =\text{MHA}(F_k, F_Q, F_V)) \oplus Concat(f_I, f_M)
		\label{eq6}
	\end{equation}
	where $Concat$ presents the concatenation operation, which is used to link $f_I$ and $f_M$, and $\oplus$ represents the element-wise summation operation.
	
	We follow the paper of \cite{vaswani2017attention} and build a multi-head attention block $\text{{MHA}}$ to implement the self-attention mechanism in the MMFA fusion module, as the effectiveness of this attention block in processing different modality data (such as sequence data \citep{vaswani2017attention} and vision data \citep{dosovitskiy2020image}) has been shown.
	\begin{equation*}
		\text{MHA}(F_K,F_Q,F_V) =  f(Concat(head_1(F_K,F_Q,F_V),...,head_h(F_K,F_Q,F_V)))
	\end{equation*}
	
	\begin{equation}
		head_i(F_K,F_Q,F_V) = \frac{Softmax((F^i_K)^\text{T} \otimes (F^i_Q))}{\sqrt{s}} \otimes v_I((F^i_V))
		\label{eq7}
	\end{equation}
	where $\sqrt{s}$ is the scaling factor, $\otimes$ is the element-wise multiplication operation and $(F^i_K)^\text{T}$ is the transpose of $F^i_K$.  
	$F^i_K \in \mathbb{R}^{d_{k}}$, $F^i_Q \in \mathbb{R}^{d_{q}}$ and  $F^i_V \in \mathbb{R}^{d_{v}}$ are the $key$, $query$ and $value$ vector in $head_i$, where $d_k = d_q = d_i = s=D_T/h$, and $h$ is the number of the heads.

	\begin{equation*}
		f_K = Concat(f_{Mk},f_{Ik})
		\label{eq6}
	\end{equation*}
	\begin{equation*}
		f_Q = Concat(f_{Mq},f_{Iq})
		\label{eq6}
	\end{equation*}
	\begin{equation}
		f_V = Concat(f_{Mv},f_{Iv})
		\label{eq6}
	\end{equation}
		
	where $k$, $q$ and $v$ indicate a kind of single-layer neural network that is a intuitive way to execute non-linear transformations on feature maps in deep learning. 
	$k_M$, $q_M$ and $v_M$ are used to transform the metadata features to have the same structure (\textit{i.e.} the same input feature number and same output feature number) but with different parameters. 
	$k_I$, $q_I$ and $v_I$ have the same structure, and they are used to transform the image features. 
			
	\begin{equation*}
		f_{Iq} , f_{Ik}, f_{Iv} := D(qkv_I(f_I)) = D(BN(f_I \otimes W_I + b_I))
	\end{equation*}
	\begin{equation}
		f_{Mq} , f_{Mk}, f_{Mv} := D(qkv_M(f_M)) = D(BN(f_M \otimes W_M + b_M))	
		\end{equation}
	
	where $W_M \in \mathbb{R}^{L_M \times d_{meta}}$ and $b_M \in \mathbb{R}^{d_{meta}}$ are the weights and biases of $k_M$, while $W_M \in \mathbb{R}^{L_I \times d_{img}}$ and $b_I \in \mathbb{R}^{d_{img}}$ are the weights and biases of $k_I$.
	$D$ means divide operation that equally dividing the output of $qkv$ into thirds, i.e., key, value and query features. 
	$BN$ indicates the batch normalization operation \citep{ioffe2015batch}.
	$F_K \in \mathbb{R}^{D_{k}}$, $F_Q \in \mathbb{R}^{D_{q}}$ and $F_V \in \mathbb{R}^{D_{v}}$ are the $query$, $key$ and $value$ feature vectors in the self-attention mechanism, where $D_k=D_q=D_v=d_{img}+d_{meta}$.
	
	$nn(x)$ is a single-layer neural network like $k$, $q$, and $v$, but with a different structure and parameters.
	$nn(x)$ is defined as:
	\begin{equation}
		nn(x) = BN(x \otimes W^{nn} + b^{nn})
		\label{eq8}
	\end{equation}
	where $W^f \in \mathbb{R}^{D_T \times L_I+L_M}$ and $b^f \in \mathbb{R}^{L_I+L_M}$ are the  weights and biases.
	\section{Experiments}
	In this section, the performance of our joint-individual fusion structure and multi-modal attention module is evaluated. 
	Five CNNs and three well-established datasets are used in our experiments.  
	The datasets, implementation details, experimental results and discussion will be introduced in sequence.
	
	\subsection{Datasets}
	Three public skin lesion classification datasets with both dermatological images and patient metadata, PAD-UFES-20 \citep{PACHECO2020106221}, Seven-Point Checklist (SPC) \citep{kawahara2019seven}, and ISIC-2019 \citep{tschandl2018ham10000, codella2018skin, combalia2019bcn20000}, are used for the performance evaluation: 
	
	\textbf{PAD-UFES-20} dataset has 2298 patient cases consisting of clinical images collected by smartphone devices and 21 metadata entries, including age, gender, skin history, parent's background and others. 
	This dataset is used to classify six-classes skin lesions: Seborrheic Keratosis (SEK), Melanoma (MEL), Nevus (NEV), Basal Cell Carcinoma (BCC), Squamous Cell Carcinoma (SCC), and Actinic Keratosis (ACK).
	
	\textbf{Seven-Point Checklist (SPC)} dataset is comprised of 413 training cases, 203 validation cases, and 395 testing cases. 
	Each case contains dermatological image data and 14 metadata entries that include gender, location of skin lesion, management, and seven-point checklist feature. 
	The SPC dataset mainly has five types of skin lesions, including MEL, NEV, SEK, BCC and Miscellaneous (MISC). 
	
	\textbf{ISIC-2019} contains 25331 dermoscopy images and each image is associated with three clinical features: location, gender and age. 
	This dataset has eight types of skin lesions: MEL, NEV, BCC, ACK, Vascular Lesion (VAL), Benign Keratosis (BK), Dermatofibroma (DF), and Squamous Cell Carcinoma (SCC). 
	%
	For the PAD-UFES-20 dataset and the ISIC-2019 dataset, we follow the paper of \cite{pacheco2021attention} and take the five-fold cross-validation stratified by the classes' frequency to evaluate our method. 
	For the SPC dataset, we used the division of the creator of training,  validation, and testing parts. 
	
		To evaluate on the SPC dataset, we train all the models for five times, and get the average and standard deviation values for the comparisons.

	\subsection{Implementation Details}
	In the experiments, we evaluated the proposed method on the three datasets and made four performance comparisons on each dataset.
	First, to show the effectiveness of patient metadata, we compared the methods using both images and metadata with those only using image data.
	Then, to demonstrate the superiority of our method, we compared the proposed JIF-MMFA method with other current fusion methods. 
	Finally, to illustrate the effectiveness of the JIF structure and the MMFA module, we separately compared the JIF structure with the Joint Fusion (JF) structure, and the MMFA module with other fusion modules.
	In the above comparisons, five CNN backbones, Mobilenet-v2 \citep{sandler2018mobilenetv2}, Efficientnet-B3 \citep{tan2019efficientnet}, Resnet-50 \citep{he2016deep}, Densenet-121 \citep{huang2017densely}, and Xception \citep{chollet2017xception}, were used as $FM_I$ to evaluate the generalization ability of the fusion methods for those CNNs. 
	Commonly-used metrics, including balanced accuracy (BAC), accuracy (ACC), and the area under the curve (AUC), were used for the performance evaluation.
	We selected the BAC value as the ranking metric \citep{pacheco2021attention}, and limited all tables in the main paper to only display the performance in terms of BAC value. 
	The other metrics can be found in the supplementary materials.
	
	All the models were first initialized with pre-trained weights of ImageNet \citep{deng2009imagenet} and then fine-tuned on the three skin cancer classification datasets for 150 epochs. 
	A SGD optimizer with an initial learning rate of 0.005 and a CosineAnnealingLR schedule in PyTorch were employed to train the models. 
	The training was stopped early when the model's validation BAC value has been not improved for 30 consecutive epochs.
	Since the dataset is imbalanced, we follow the paper of \cite{pacheco2021attention} and used the class-weighted cross-entropy as the loss function. 
	In our JIF structure, there are three-branch loss functions: $L_I$, $L_M$ and $L_{IM}$.  
	In the training process, the goal of the JIF structure is minimizing the total loss function $L_{total} = \beta*L_I + (1-\beta)*L_M  + L_{IM}$, where $\beta$ is the weight of each modality in the whole training.
	We set $\beta$ to 0.5, as we consider the two-modality data equally important.
	All the images were resized to 224 $\times$ 224 $\times$ 3 before the training, and common data augmentations, including horizontal and vertical flipping, shifting, rotation and scaling, were used to expand the datasets.
	The Python libraries Pytorch \citep{paszke2019pytorch}, Sklearn \citep{scikit-learn}, Numpy \citep{harris2020array}, and Albumentation \citep{buslaev2020albumentations}, were used to build our workflow, including model design, data loader, training and testing flows.

	\subsection{Experiments results}
	To simplify the description in the following text, we use the following abbreviations: \textbf{JF:} Joint Fusion; \textbf{JIF:} Joint-Individual Fusion; \textbf{OFB:} Only fusion branch; \textbf{FS:} Fusion structure; \textbf{Cat:} Concatenation; \textbf{MB:} Metablock; \textbf{MN:} MetaNet; \textbf{MMFA:} Multi-Modal Fusion Attention. 
	Also, we concatenate these abbreviations to name the employed fusion methods. 
	For example, we abbreviate our Joint-Individual Fusion (JIF) structure with multi-modal fusion attention (MMFA) as \textbf{JIF-MMFA}, and the Joint Fusion structure (JF) with metablock (MB) as \textbf{JF-MB}.
	Additionally, \textbf{JIF-MMFA (OFB)} denotes the result only from the $P_{IM}$ of the JIF-MMFA method; 
	while \textbf{JIF-MMFA (All)} denotes the result by averaging the three predictions $P_I$, $P_M$, and $P_{IM}$ of the JIF-MMFA method (see Fig.~\ref{fig1}(b)).
	
	In Table~\ref{table1}, Table~\ref{table2}, and Table~\ref{table3}, we show the performance comparisons between our method and other currently existing methods, according to the mean value and standard deviation of the BAC metric. 
	These comparisons are used to show the effectiveness of using the metadata, the superiority of the proposed method, and ablation studies of our JIF-MMFA.
	Table~\ref{table4}, Table~\ref{table5}, and Table~\ref{table6}, display the Wilcoxon test of the methods as shown in Table~\ref{table1}, Table~\ref{table2}, and Table~\ref{table3}, respectively, to further compare these methods in terms of their statistical difference.
	Table~\ref{table7}, Table~\ref{table8}, and Table~\ref{table9} separately present the experimental results of the JIF structure and the JF structure with different fusion modules, which analyzes the effectiveness of the JIF structure and the MMFA Module.

	\begin{table}[thp]
	\centering
	\setlength{\abovecaptionskip}{0pt}
	\setlength{\belowcaptionskip}{10pt}
	
	\caption{Performance comparisons between JIF-MMFA and other methods on the PAD-UFES-20 dataset in terms of BAC.  $FS$: Fusion structure; $Cat$: Concatenation; $MB$: Metablock; $MN$: MetaNet; $MA$: Mutual Attention, $MMFA$ Multi-Modal Fusion Attention.  $JF$: Joint Fusion Structure, $JIF$: Joint-Individual Fusion Structure.	}

	\scalebox{0.65}{
		\renewcommand\arraystretch{1.75}
		\setlength{\tabcolsep}{1mm}{

	\begin{tabular}{|c|c|ccccc|cc|}
		\hline
		\textit{\textbf{FS}}        & \multirow{2}{*}{\textit{\textbf{Image}}} & \multicolumn{5}{c|}{\textit{\textbf{JF}}}                                                                                                                                                               & \multicolumn{2}{c|}{\textit{\textbf{JIF}}}                                     \\ \cline{1-1} \cline{3-9} 
		\textit{\textbf{CNN}}       &                                          & \multicolumn{1}{c|}{\textit{\textbf{CAT}}} & \multicolumn{1}{c|}{\textit{\textbf{MB}}} & \multicolumn{1}{c|}{\textit{\textbf{MN}}} & \multicolumn{1}{c|}{\textit{\textbf{MA}}} & \textit{\textbf{MMFA}} & \multicolumn{1}{c|}{\textit{\textbf{MMFA(OFB)}}} & \textit{\textbf{MMFA(All)}} \\ \hline
		\textit{\textbf{densenet}}  & 68.9$\pm$2.6                                & \multicolumn{1}{c|}{73.8$\pm$1.4}             & \multicolumn{1}{c|}{72.4$\pm$2.1}            & \multicolumn{1}{c|}{68.6$\pm$2.2}            & \multicolumn{1}{c|}{76.0$\pm$2.3}            & 75.6$\pm$1.7              & \multicolumn{1}{c|}{\textbf{78.0$\pm$2.0}}          & 77.7$\pm$1.8                   \\ \hline
		\textit{\textbf{mobilenet}} & 67.1$\pm$1.5                                & \multicolumn{1}{c|}{73.7$\pm$1.2}             & \multicolumn{1}{c|}{70.1$\pm$3.7}            & \multicolumn{1}{c|}{69.1$\pm$3.0}            & \multicolumn{1}{c|}{75.0$\pm$1.6}            & 75.2$\pm$1.6              & \multicolumn{1}{c|}{74.7$\pm$1.4}                   & \textbf{75.6$\pm$0.7}          \\ \hline
		\textit{\textbf{resnet}}    & 66.1$\pm$1.5                                & \multicolumn{1}{c|}{72.9$\pm$1.7}             & \multicolumn{1}{c|}{72.1$\pm$1.6}            & \multicolumn{1}{c|}{68.8$\pm$3.0}            & \multicolumn{1}{c|}{73.3$\pm$1.6}            & 73.6$\pm$2.4              & \multicolumn{1}{c|}{76.0$\pm$1.2}                   & \textbf{76.4$\pm$1.5}          \\ \hline
		\textit{\textbf{effnet}}    & 64.6$\pm$1.4                                & \multicolumn{1}{c|}{76.8$\pm$1.4}             & \multicolumn{1}{c|}{71.4$\pm$2.2}            & \multicolumn{1}{c|}{65.4$\pm$2.0}            & \multicolumn{1}{c|}{74.8$\pm$2.0}            & 76.0$\pm$1.8              & \multicolumn{1}{c|}{78.8$\pm$1.6}                   & \textbf{79.8$\pm$1.4}          \\ \hline
		\textit{\textbf{xception}}  & 68.3$\pm$1.5                                & \multicolumn{1}{c|}{73.8$\pm$1.9}             & \multicolumn{1}{c|}{70.1$\pm$1.6}            & \multicolumn{1}{c|}{66.8$\pm$1.3}            & \multicolumn{1}{c|}{73.5$\pm$2.9}            & 74.1$\pm$3.0              & \multicolumn{1}{c|}{75.9$\pm$1.4}                   & \textbf{76.3$\pm$1.2}          \\ \hline
		\textit{\textbf{Average}}   & 67.0$\pm$2.3                                & \multicolumn{1}{c|}{74.2$\pm$2.0}             & \multicolumn{1}{c|}{71.2$\pm$2.6}            & \multicolumn{1}{c|}{67.8$\pm$2.8}            & \multicolumn{1}{c|}{74.5$\pm$2.4}            & 74.9$\pm$2.3              & \multicolumn{1}{c|}{76.7$\pm$2.2}                   & \textbf{77.2$\pm$2.0}          \\ \hline
	\end{tabular}

	}}
	\label{table1}
\end{table}	
	
	\begin{table}[thp]
		\centering
		\setlength{\abovecaptionskip}{0pt}
		\setlength{\belowcaptionskip}{10pt}
				
		\caption{Performance comparisons between JIF-MMFA and other methods on the SPC dataset in terms of BAC.  $FS$: Fusion structure; $Cat$: Concatenation; $MB$: Metablock; $MN$: MetaNet; $MA$: Mutual Attention, $MMFA$ Multi-Modal Fusion Attention.  $JF$: Joint Fusion Structure, $JIF$: Joint-Individual Fusion Structure.}
				
		\scalebox{0.65}{
			\renewcommand\arraystretch{1.75}
			\setlength{\tabcolsep}{1mm}{

	\begin{tabular}{|c|c|ccccc|cc|}
		\hline
		\textit{\textbf{FS}}        & \multirow{2}{*}{\textit{\textbf{Image}}} & \multicolumn{5}{c|}{\textit{\textbf{JF}}}                                                                                                                                                               & \multicolumn{2}{c|}{\textit{\textbf{JIF}}}                                     \\ \cline{1-1} \cline{3-9} 
		\textit{\textbf{CNN}}       &                                          & \multicolumn{1}{c|}{\textit{\textbf{CAT}}} & \multicolumn{1}{c|}{\textit{\textbf{MB}}} & \multicolumn{1}{c|}{\textit{\textbf{MN}}} & \multicolumn{1}{c|}{\textit{\textbf{MA}}} & \textit{\textbf{MMFA}} & \multicolumn{1}{c|}{\textit{\textbf{MMFA(OFB)}}} & \textit{\textbf{MMFA(All)}} \\ \hline
		\textit{\textbf{densenet}}  & 54.9$\pm$2.9                                & \multicolumn{1}{c|}{61.1$\pm$2.2}             & \multicolumn{1}{c|}{67.4$\pm$0.6}            & \multicolumn{1}{c|}{57.5$\pm$1.9}            & \multicolumn{1}{c|}{69.5$\pm$3.3}            & 72.3$\pm$2.6              & \multicolumn{1}{c|}{70.9$\pm$2.3}                   & \textbf{73.1$\pm$2.6}          \\ \hline
		\textit{\textbf{mobilenet}} & 57.4$\pm$4.8                                & \multicolumn{1}{c|}{70.3$\pm$1.2}             & \multicolumn{1}{c|}{69.3$\pm$0.9}            & \multicolumn{1}{c|}{60.2$\pm$4.0}            & \multicolumn{1}{c|}{70.4$\pm$0.9}            & 69.4$\pm$3.7              & \multicolumn{1}{c|}{72.1$\pm$4.9}                   & \textbf{73.1$\pm$3.9}          \\ \hline
		\textit{\textbf{resnet}}    & 53.7$\pm$4.2                                & \multicolumn{1}{c|}{62.8$\pm$5.1}             & \multicolumn{1}{c|}{67.8$\pm$1.8}            & \multicolumn{1}{c|}{55.0$\pm$2.2}            & \multicolumn{1}{c|}{65.7$\pm$3.0}            & 67.5$\pm$2.6              & \multicolumn{1}{c|}{70.0$\pm$2.7}                   & \textbf{70.4$\pm$2.6}          \\ \hline
		\textit{\textbf{effnet}}    & 55.0$\pm$1.4                                & \multicolumn{1}{c|}{73.2$\pm$2.3}             & \multicolumn{1}{c|}{68.2$\pm$2.3}            & \multicolumn{1}{c|}{55.1$\pm$2.4}            & \multicolumn{1}{c|}{70.0$\pm$1.9}            & 70.8$\pm$1.2              & \multicolumn{1}{c|}{71.2$\pm$2.0}                   & \textbf{74.0$\pm$1.1}          \\ \hline
		\textit{\textbf{xception}}  & 55.7$\pm$3.7                                & \multicolumn{1}{c|}{\textbf{72.8$\pm$2.2}}    & \multicolumn{1}{c|}{67.0$\pm$1.4}            & \multicolumn{1}{c|}{57.4$\pm$3.0}            & \multicolumn{1}{c|}{68.9$\pm$3.4}            & 68.1$\pm$2.8              & \multicolumn{1}{c|}{70.6$\pm$2.0}                   & 71.5$\pm$2.7                   \\ \hline
		\textit{\textbf{Average}}   & 55.4$\pm$3.8                                & \multicolumn{1}{c|}{68.1$\pm$5.9}             & \multicolumn{1}{c|}{68.0$\pm$1.7}            & \multicolumn{1}{c|}{57.1$\pm$3.4}            & \multicolumn{1}{c|}{68.9$\pm$3.1}            & 69.6$\pm$3.2              & \multicolumn{1}{c|}{71.0$\pm$3.1}                   & \textbf{72.4$\pm$3.0}          \\ \hline
	\end{tabular}

		}}
		\label{table2}
	\end{table}
	
	\begin{table}[thp]
		\centering
		\setlength{\abovecaptionskip}{0pt}
		\setlength{\belowcaptionskip}{10pt}

		\caption{Performance comparisons between JIF-MMFA and other methods on the ISIC-2019 dataset in terms of BAC. $FS$: Fusion structure; $Cat$: Concatenation; $MB$: Metablock; $MN$: MetaNet; $MA$: Mutual Attention, $MMFA$ Multi-Modal Fusion Attention.  $JF$: Joint Fusion Structure, $JIF$: Joint-Individual Fusion Structure.
		}
		
		\scalebox{0.65}{
			\renewcommand\arraystretch{1.75}
			\setlength{\tabcolsep}{1mm}{

	\begin{tabular}{|c|c|ccccc|cc|}
		\hline
		\textit{\textbf{FS}}        & \multirow{2}{*}{\textit{\textbf{Image}}} & \multicolumn{5}{c|}{\textit{\textbf{JF}}}                                                                                                                                                               & \multicolumn{2}{c|}{\textit{\textbf{JIF}}}                                     \\ \cline{1-1} \cline{3-9} 
		\textit{\textbf{CNN}}       &                                          & \multicolumn{1}{c|}{\textit{\textbf{CAT}}} & \multicolumn{1}{c|}{\textit{\textbf{MB}}} & \multicolumn{1}{c|}{\textit{\textbf{MN}}} & \multicolumn{1}{c|}{\textit{\textbf{MA}}} & \textit{\textbf{MMFA}} & \multicolumn{1}{c|}{\textit{\textbf{MMFA(OFB)}}} & \textit{\textbf{MMFA(All)}} \\ \hline
		\textit{\textbf{densenet}}  & 81.8$\pm$0.5                                & \multicolumn{1}{c|}{83.3$\pm$1.0}             & \multicolumn{1}{c|}{82.9$\pm$0.5}            & \multicolumn{1}{c|}{82.9$\pm$1.6}            & \multicolumn{1}{c|}{82.4$\pm$1.1}            & 82.4$\pm$0.7              & \multicolumn{1}{c|}{\textbf{84.8$\pm$1.1}}          & 84.6$\pm$0.9                   \\ \hline
		\textit{\textbf{mobilenet}} & 80.3$\pm$1.7                                & \multicolumn{1}{c|}{83.0$\pm$0.7}             & \multicolumn{1}{c|}{82.9$\pm$1.0}            & \multicolumn{1}{c|}{83.4$\pm$0.2}            & \multicolumn{1}{c|}{81.8$\pm$0.6}            & 81.6$\pm$1.2              & \multicolumn{1}{c|}{\textbf{85.0$\pm$1.5}}          & 84.8$\pm$1.4                   \\ \hline
		\textit{\textbf{resnet}}    & 81.5$\pm$0.4                                & \multicolumn{1}{c|}{82.7$\pm$1.1}             & \multicolumn{1}{c|}{83.4$\pm$0.4}            & \multicolumn{1}{c|}{83.4$\pm$0.8}            & \multicolumn{1}{c|}{65.5$\pm$9.4}            & 68.8$\pm$5.3              & \multicolumn{1}{c|}{83.7$\pm$0.5}                   & \textbf{83.7$\pm$0.3}          \\ \hline
		\textit{\textbf{effnet}}    & 79.4$\pm$0.7                                & \multicolumn{1}{c|}{80.2$\pm$0.5}             & \multicolumn{1}{c|}{79.3$\pm$1.7}            & \multicolumn{1}{c|}{79.6$\pm$0.7}            & \multicolumn{1}{c|}{81.9$\pm$1.2}            & 80.8$\pm$1.5              & \multicolumn{1}{c|}{\textbf{82.6$\pm$0.6}}          & 82.5$\pm$0.7                   \\ \hline
		\textit{\textbf{xception}}  & 79.2$\pm$1.4                                & \multicolumn{1}{c|}{79.8$\pm$0.9}             & \multicolumn{1}{c|}{78.2$\pm$0.6}            & \multicolumn{1}{c|}{79.0$\pm$0.4}            & \multicolumn{1}{c|}{82.1$\pm$1.4}            & 81.1$\pm$1.5              & \multicolumn{1}{c|}{82.5$\pm$0.3}                   & \textbf{82.7$\pm$0.3}          \\ \hline
		\textit{\textbf{Average}}   & 80.4$\pm$1.5                                & \multicolumn{1}{c|}{81.8$\pm$1.7}             & \multicolumn{1}{c|}{81.3$\pm$2.3}            & \multicolumn{1}{c|}{81.7$\pm$2.1}            & \multicolumn{1}{c|}{78.7$\pm$7.9}            & 78.9$\pm$5.8              & \multicolumn{1}{c|}{\textbf{83.8$\pm$1.4}}          & 83.7$\pm$1.3                   \\ \hline
	\end{tabular}
		}}
		\label{table3}
	\end{table}
	
	\begin{table}[thp]
		\centering
		\setlength{\abovecaptionskip}{0pt}
		\setlength{\belowcaptionskip}{10pt}
		
		\caption{The results of the statistical test (Wilcoxon pair-wise test) for all the methods on the PAD-UFES-20 dataset. 
			The $P_{value}>0.05$ is highlighted in bold.
		}
		
		\scalebox{0.55}{
			\renewcommand\arraystretch{1.75}
			\setlength{\tabcolsep}{1.mm}{

	\begin{tabular}{|c|c|c|c|}
		\hline
		\textit{\textbf{Model-Pairs}} & \textit{\textbf{P\_value}}    & \textit{\textbf{Model-Pairs}}   & \textit{\textbf{P\_value}}    \\ \hline
		Image - JF-CAT                & 5.96E-08                      & JF-MB - JF-MA                   & 1.01E-05                      \\ \hline
		Image - JF-MB                 & 2.56E-06                      & JF-MB - JF-MMFA                 & 1.13E-06                      \\ \hline
		Image - JF-MN                 & \textbf{0.2635}          & JF-MB - JIF-MMFA (OFB)          & 1.19E-07                      \\ \hline
		Image - JF-MA                 & 1.19E-07                      & JF-MB - JIF-MMFA (All)          & 1.19E-07                      \\ \hline
		Image - JF-MMFA               & 1.19E-07                      & JF-MN - JF-MA                   & 5.96E-08                      \\ \hline
		Image - JIF-MMFA (OFB)        & 5.96E-08                      & JF-MN - JF-MMFA                 & 1.79E-07                      \\ \hline
		Image - JIF-MMFA (All)        & 5.96E-08                      & JF-MN - JIF-MMFA (OFB)          & 1.19E-07                      \\ \hline
		JF-CAT - JF-MB                & 6.37E-05                      & JF-MN - JIF-MMFA (All)          & 5.96E-08                      \\ \hline
		JF-CAT - JF-MN                & 1.19E-07                      & JF-MA - JF-MMFA                 & \textit{\textbf{0.5249}} \\ \hline
		JF-CAT - JF-MA                & \textit{\textbf{0.3957}} & JF-MA - JIF-MMFA (OFB)          & 0.000714958                   \\ \hline
		JF-CAT - JF-MMFA              & \textit{\textbf{0.2411}} & JF-MA - JIF-MMFA (All)          & 2.21E-05                      \\ \hline
		JF-CAT - JIF-MMFA (OFB)       & 2.21E-05                      & JF-MMFA - JIF-MMFA (OFB)        & 0.004175186                   \\ \hline
		JF-CAT - JIF-MMFA (All)       & 1.19E-07                      & JF-MMFA - JIF-MMFA (All)        & 0.000216901                   \\ \hline
		JF-MB - JF-MN                 & 8.80E-05                      & JIF-MMFA (OFB) - JIF-MMFA (All) & 0.001815677                   \\ \hline
	\end{tabular}}}

		\label{table4}
	\end{table}

	\begin{table}[thp]
		\centering
		\setlength{\abovecaptionskip}{0pt}
		\setlength{\belowcaptionskip}{10pt}
		
		\caption{The results of the statistical test (Wilcoxon pair-wise test) for all the methods on the SPC dataset. 
			The $P_{value}>0.05$ is highlighted in bold.
		}
		
		\scalebox{0.55}{
			\renewcommand\arraystretch{1.75}
			\setlength{\tabcolsep}{1.mm}{

	\begin{tabular}{|c|c|c|c|}
		\hline
		\textit{\textbf{Model-Pairs}} & \textit{\textbf{P\_value}} & \textit{\textbf{Model-Pairs}}   & \textit{\textbf{P\_value}} \\ \hline
		Image - JF-CAT                & 4.17E-07                   & JF-MB - JF-MA                   & \textbf{0.1073}            \\ \hline
		Image - JF-MB                 & 5.96E-08                   & JF-MB - JF-MMFA                 & 0.0236                     \\ \hline
		Image - JF-MN                 & \textbf{0.1135}            & JF-MB - JIF-MMFA (OFB)          & 1.83E-05                   \\ \hline
		Image - JF-MA                 & 5.96E-08                   & JF-MB - JIF-MMFA (All)          & 1.79E-07                   \\ \hline
		Image - JF-MMFA               & 5.96E-08                   & JF-MN - JF-MA                   & 5.96E-08                   \\ \hline
		Image - JIF-MMFA (OFB)        & 5.96E-08                   & JF-MN - JF-MMFA                 & 5.96E-08                   \\ \hline
		Image - JIF-MMFA (All)        & 5.96E-08                   & JF-MN - JIF-MMFA (OFB)          & 5.96E-08                   \\ \hline
		JF-CAT - JF-MB                & 0.832509398                & JF-MN - JIF-MMFA (All)          & 5.96E-08                   \\ \hline
		JF-CAT - JF-MN                & 1.79E-07                   & JF-MA - JF-MMFA                 & \textit{\textbf{0.3123}}   \\ \hline
		JF-CAT - JF-MA                & \textit{\textbf{0.6915}}   & JF-MA - JIF-MMFA (OFB)          & 0.0088                     \\ \hline
		JF-CAT - JF-MMFA              & \textit{\textbf{0.4578}}   & JF-MA - JIF-MMFA (All)          & 1.23E-05                   \\ \hline
		JF-CAT - JIF-MMFA (OFB)       & 0.0255                     & JF-MMFA - JIF-MMFA (OFB)        & \textbf{0.1073}            \\ \hline
		JF-CAT - JIF-MMFA (All)       & 0.0022                     & JF-MMFA - JIF-MMFA (All)        & 7.50E-05                   \\ \hline
		JF-MB - JF-MN                 & 5.96E-08                   & JIF-MMFA (OFB) - JIF-MMFA (All) & 4.54E-05                   \\ \hline
	\end{tabular}

		}}
		\label{table5}
	\end{table}

	\begin{table}[thp]
		\centering
		\setlength{\abovecaptionskip}{0pt}
		\setlength{\belowcaptionskip}{10pt}
		
		\caption{The results of the statistical test (Wilcoxon pair-wise test) for all the methods on the ISIC-2019 dataset. 
			The $P_{value}>0.05$ is highlighted in bold.
		}
		
		\scalebox{0.55}{
			\renewcommand\arraystretch{1.75}
			\setlength{\tabcolsep}{1.mm}{

	\begin{tabular}{|c|c|c|c|}
		\hline
		\textit{\textbf{Model-Pairs}} & \textit{\textbf{P\_value}} & \textit{\textbf{Model-Pairs}}   & \textit{\textbf{P\_value}}    \\ \hline
		Image - JF-CAT                & 6.56E-06                   & JF-MB - JF-MA                   & \textit{\textbf{0.8119}}      \\ \hline
		Image - JF-MB                 & 0.0309                     & JF-MB - JF-MMFA                 & \textit{\textbf{0.4742}}      \\ \hline
		Image - JF-MN                 & 0.0025                     & JF-MB - JIF-MMFA (OFB)          & 2.56E-06                      \\ \hline
		Image - JF-MA                 & \textit{\textbf{0.3254}}   & JF-MB - JIF-MMFA (All)          & 3.28E-06                      \\ \hline
		Image - JF-MMFA               & \textit{\textbf{0.7712}}   & JF-MN - JF-MA                   & \textit{\textbf{0.6528}}      \\ \hline
		Image - JIF-MMFA (OFB)        & 5.96E-08                   & JF-MN - JF-MMFA                 & \textit{\textbf{0.2411}}      \\ \hline
		Image - JIF-MMFA (All)        & 5.96E-08                   & JF-MN - JIF-MMFA (OFB)          & 1.23E-05                      \\ \hline
		JF-CAT - JF-MB                & \textit{\textbf{0.1730}}   & JF-MN - JIF-MMFA (All)          & 1.23E-05                      \\ \hline
		JF-CAT - JF-MN                & \textit{\textbf{0.4578}}   & JF-MA - JF-MMFA                 & \textit{\textbf{0.6915}} \\ \hline
		JF-CAT - JF-MA                & \textit{\textbf{0.4108}}   & JF-MA - JIF-MMFA (OFB)          & 5.39E-05                      \\ \hline
		JF-CAT - JF-MMFA              & \textit{\textbf{0.0957}}   & JF-MA - JIF-MMFA (All)          & 1.83E-05                      \\ \hline
		JF-CAT - JIF-MMFA (OFB)       & 5.25E-06                   & JF-MMFA - JIF-MMFA (OFB)        & 1.13E-06             \\ \hline
		JF-CAT - JIF-MMFA (All)       & 2.56E-06                   & JF-MMFA - JIF-MMFA (All)        & 5.96E-07                      \\ \hline
		JF-MB - JF-MN                 & \textit{\textbf{0.1485}}   & JIF-MMFA (OFB) - JIF-MMFA (All) & \textit{\textbf{0.6073}}      \\ \hline
	\end{tabular}

		}}
		\label{table6}
	\end{table}
	
	\begin{figure}[]
		\centering
		\subfigure[Image]{\includegraphics[width=4.2cm, height=5cm]{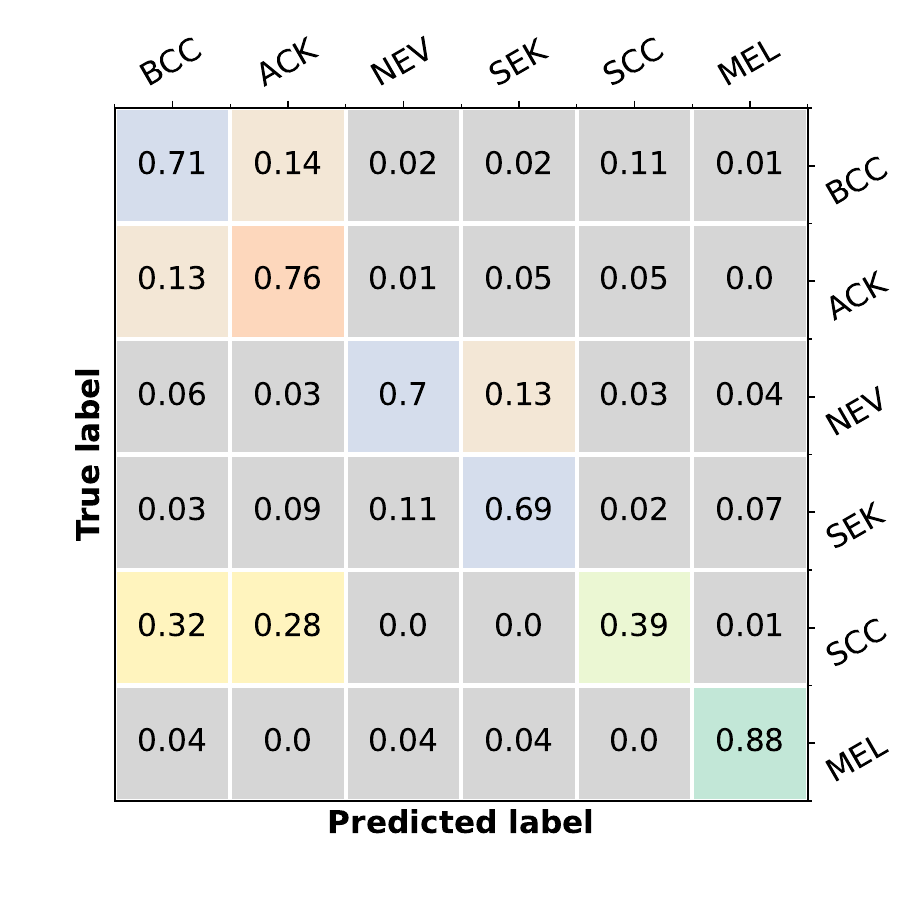}} 
		\subfigure[JF-CAT]{\includegraphics[width=4.2cm, height=5cm]{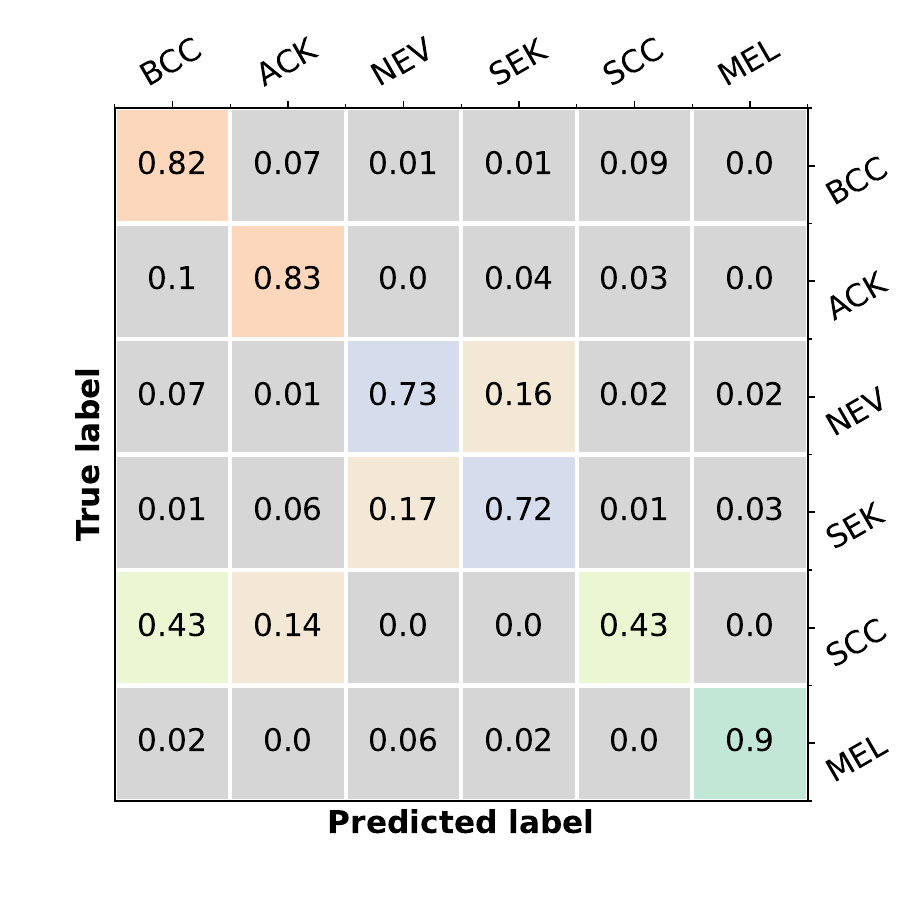}}
		\subfigure[JF-MB]{\includegraphics[width=4.2cm, height=5cm]{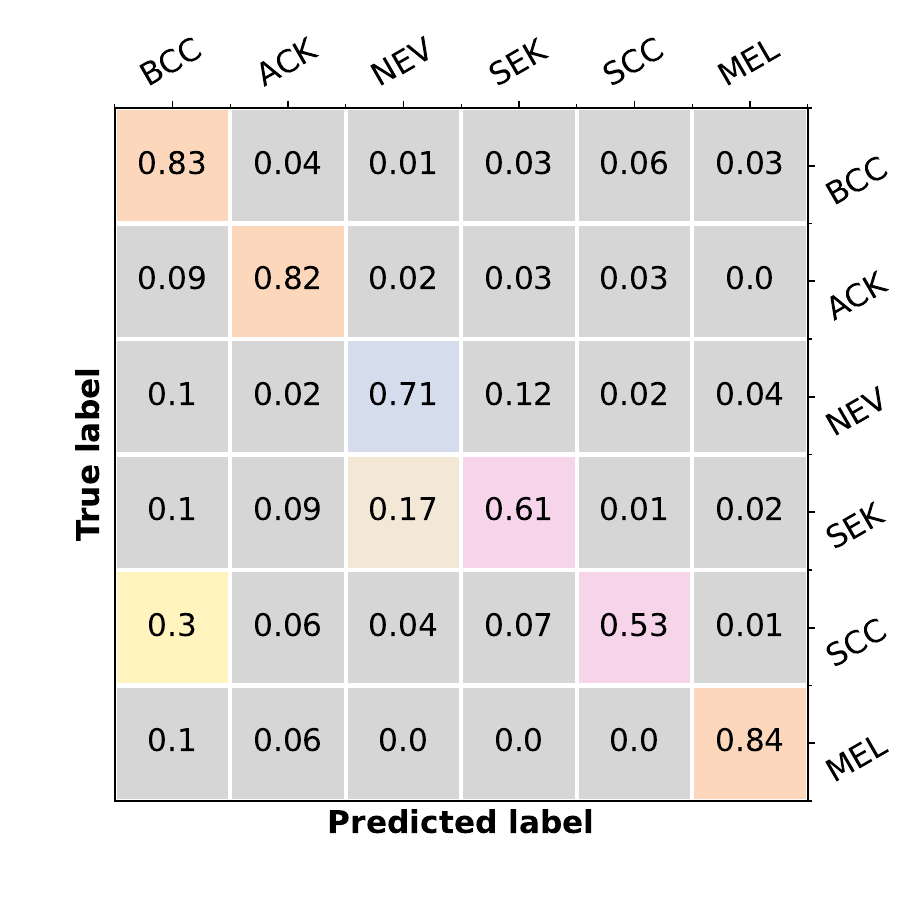}} 
		\\ 
		\centering
		\subfigure[JF-MN]{\includegraphics[width=4.2cm, height=5cm]{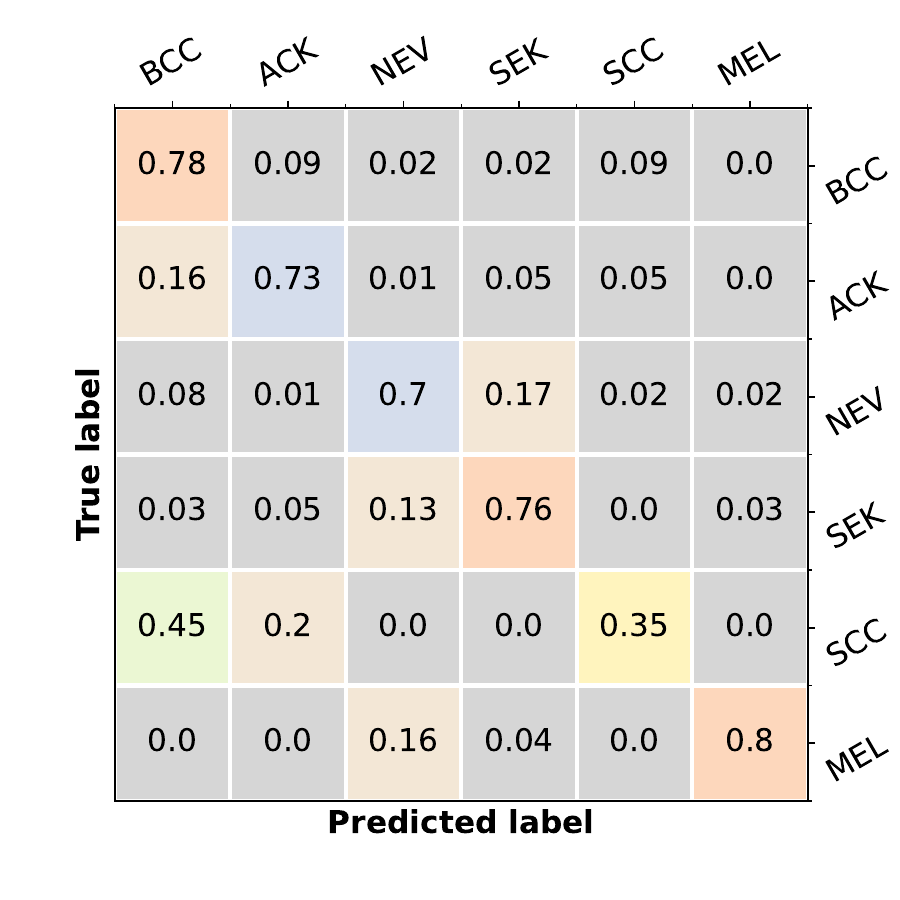}}
		\subfigure[JF-MMFA]{\includegraphics[width=4.2cm, height=5cm]{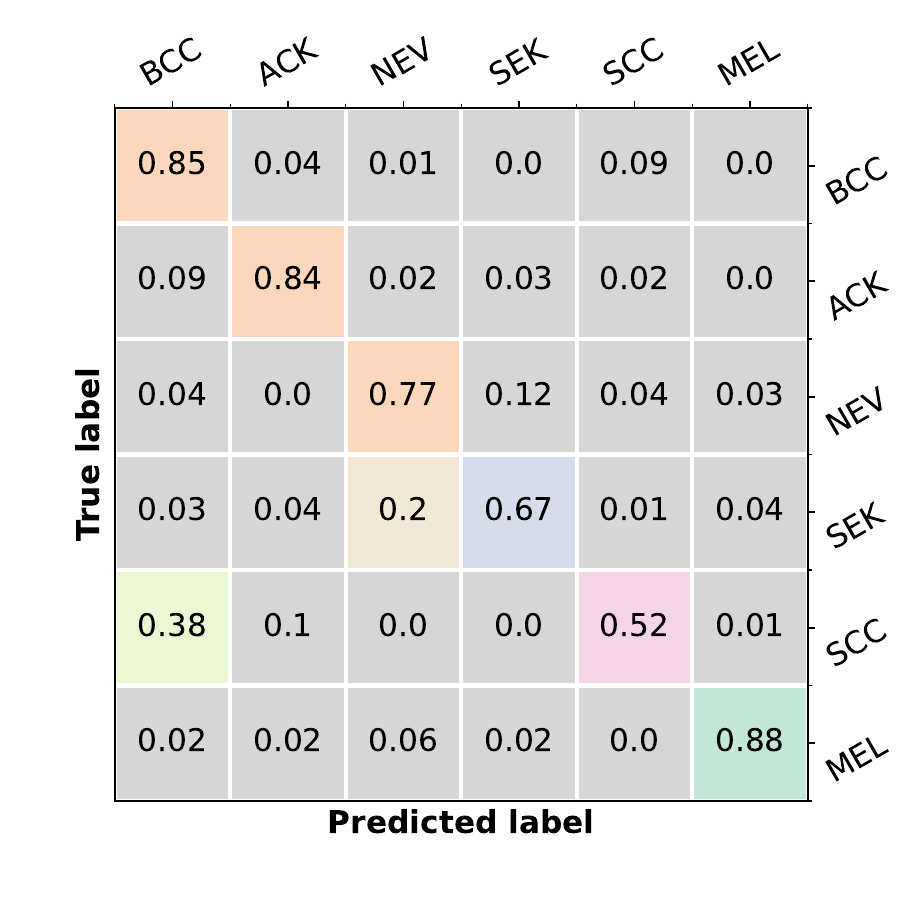}}
		\subfigure[JF-MA]{\includegraphics[width=4.2cm, height=5cm]{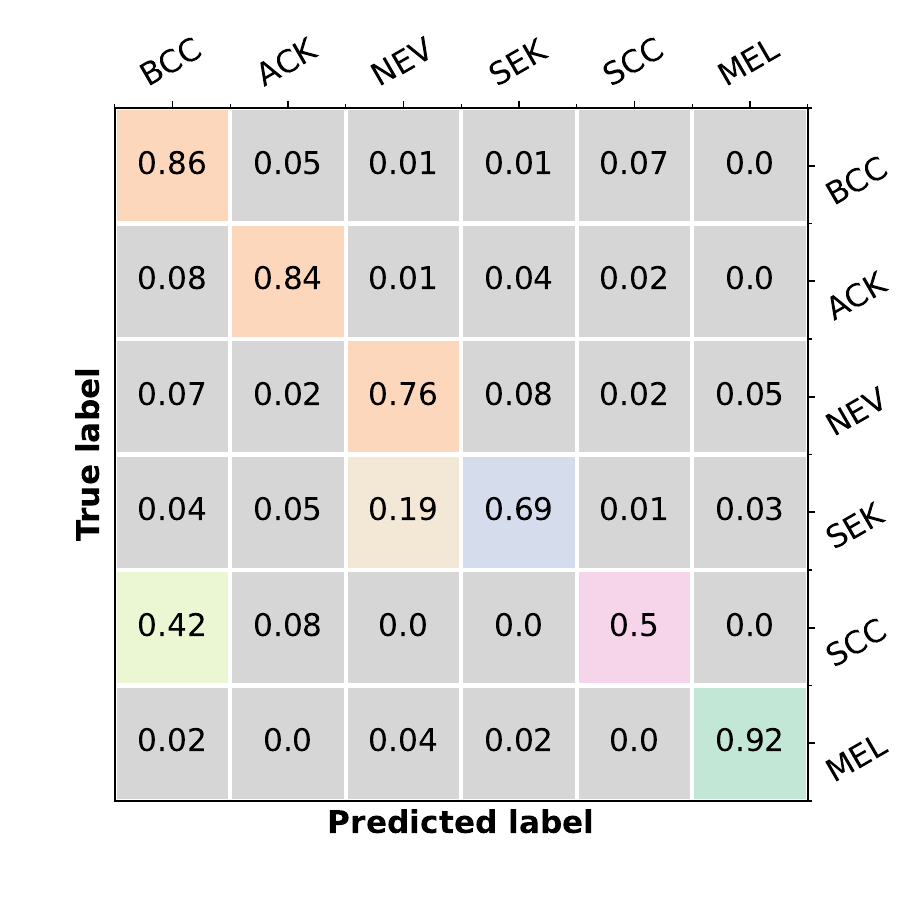}}		
		
		\subfigure[JIF-MMFA(OFB)]{\includegraphics[width=4.2cm, height=5cm]{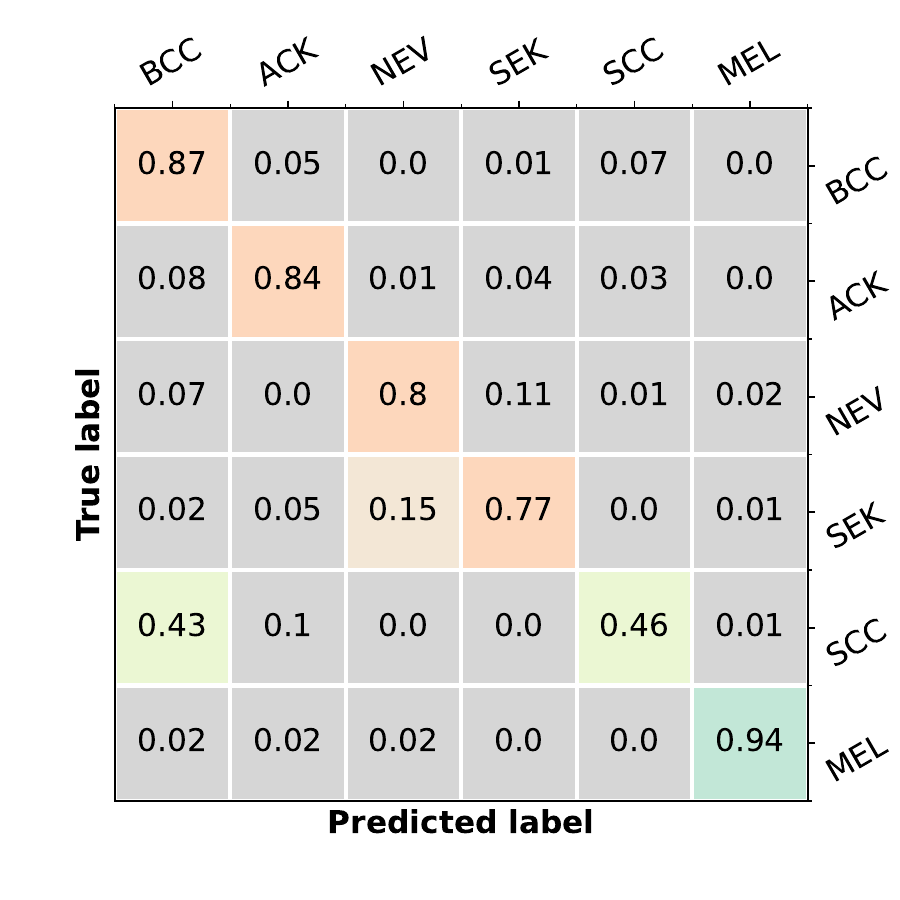}}		
		\subfigure[JIF-MMFA(All)]{\includegraphics[width=4.2cm, height=5cm]{./figures/new_fig.5/heatmaps/JointFusion_FA_densenet_averaged_heatmap_pred.pdf}}
				
		\caption{\textcolor{black}{The confusion matrix of the methods in Table~\ref{table1} considering DenseNet-121 on the PAD-UFES-20 dataset. 
				BCC: Basal Cell Carcinoma; ACK: Actinic Keratosis; NEV: Nevus; SEK: Seborrheic Keratosis; MEL: Melanoma;  SCC: Squamous Cell Carcinoma. See also sections 3.3.2 and 4.2.}} 
		
		\vspace{-0.2in}
		\label{fig4}
	\end{figure}

	\begin{figure}[]
		\centering
		\subfigure[Image]{\includegraphics[width=4.2cm, height=5cm]{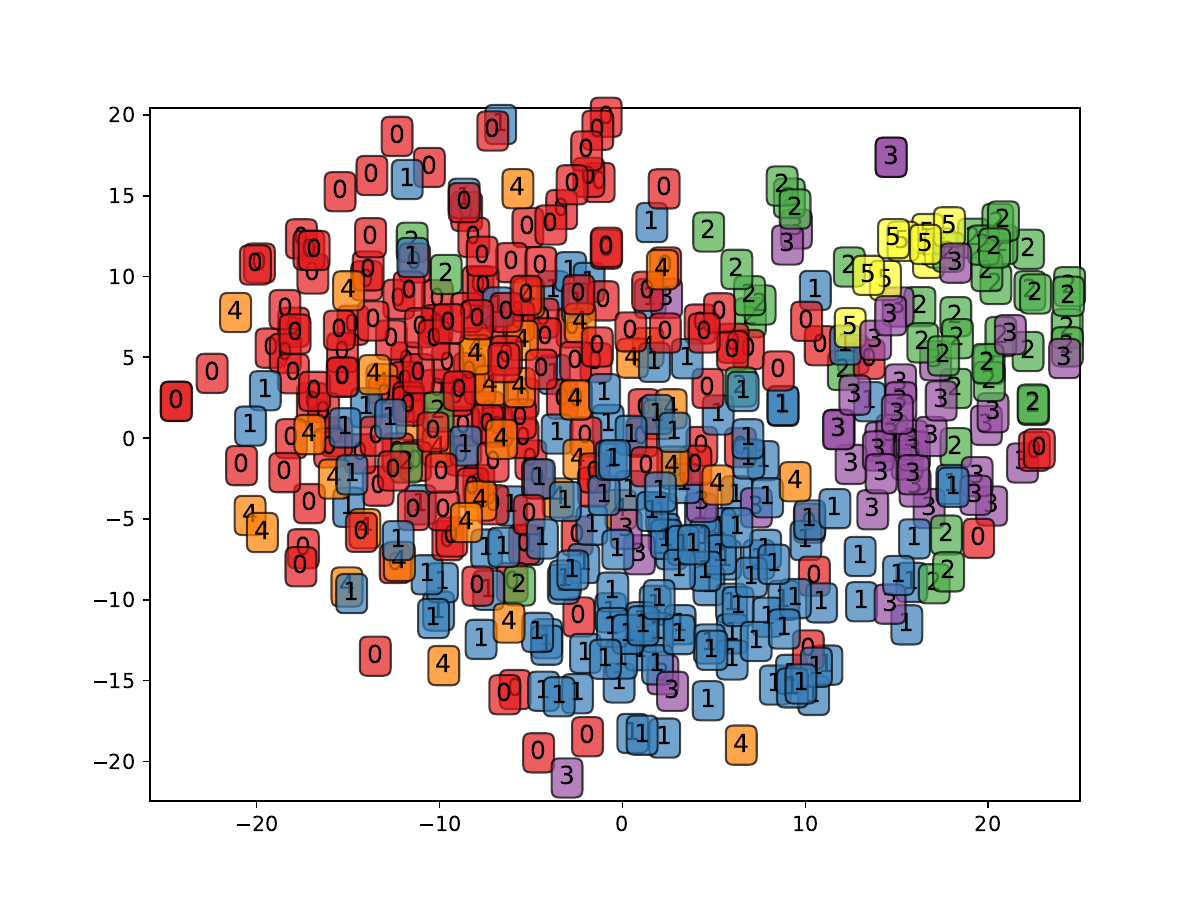}} 
		\subfigure[JF-CAT]{\includegraphics[width=4.2cm, height=5cm]{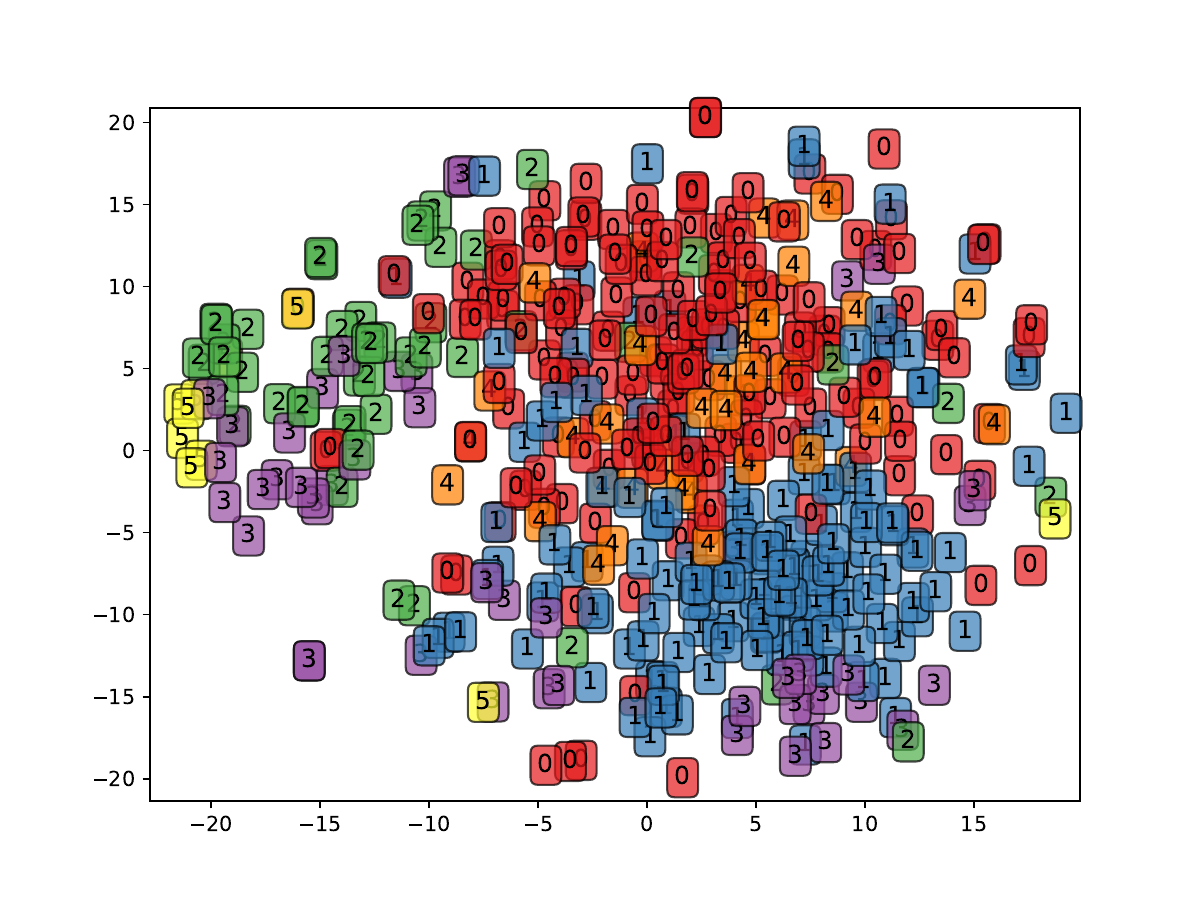}}
		\subfigure[JF-MB]{\includegraphics[width=4.2cm, height=5cm]{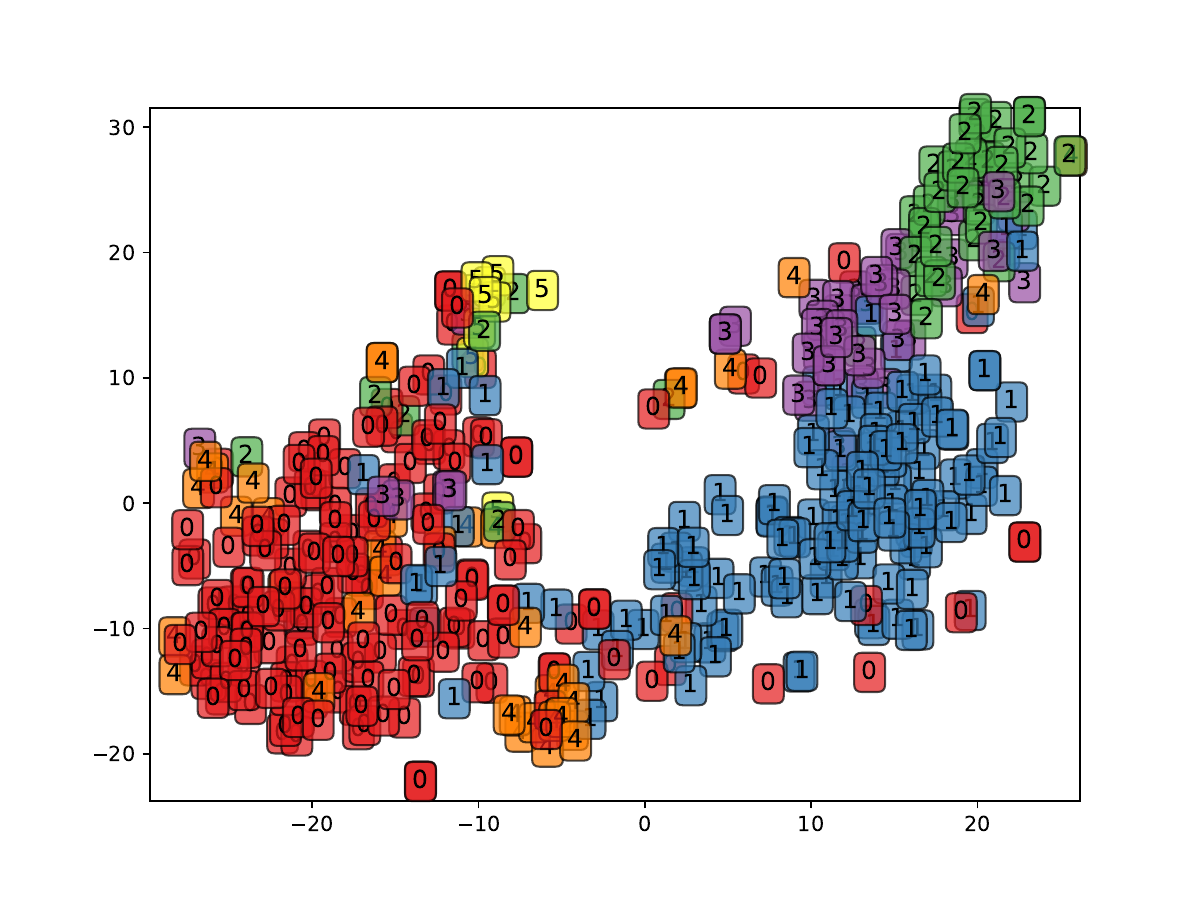}} 
		\\ 
		\centering
		\subfigure[JF-MN]{\includegraphics[width=4.2cm, height=5cm]{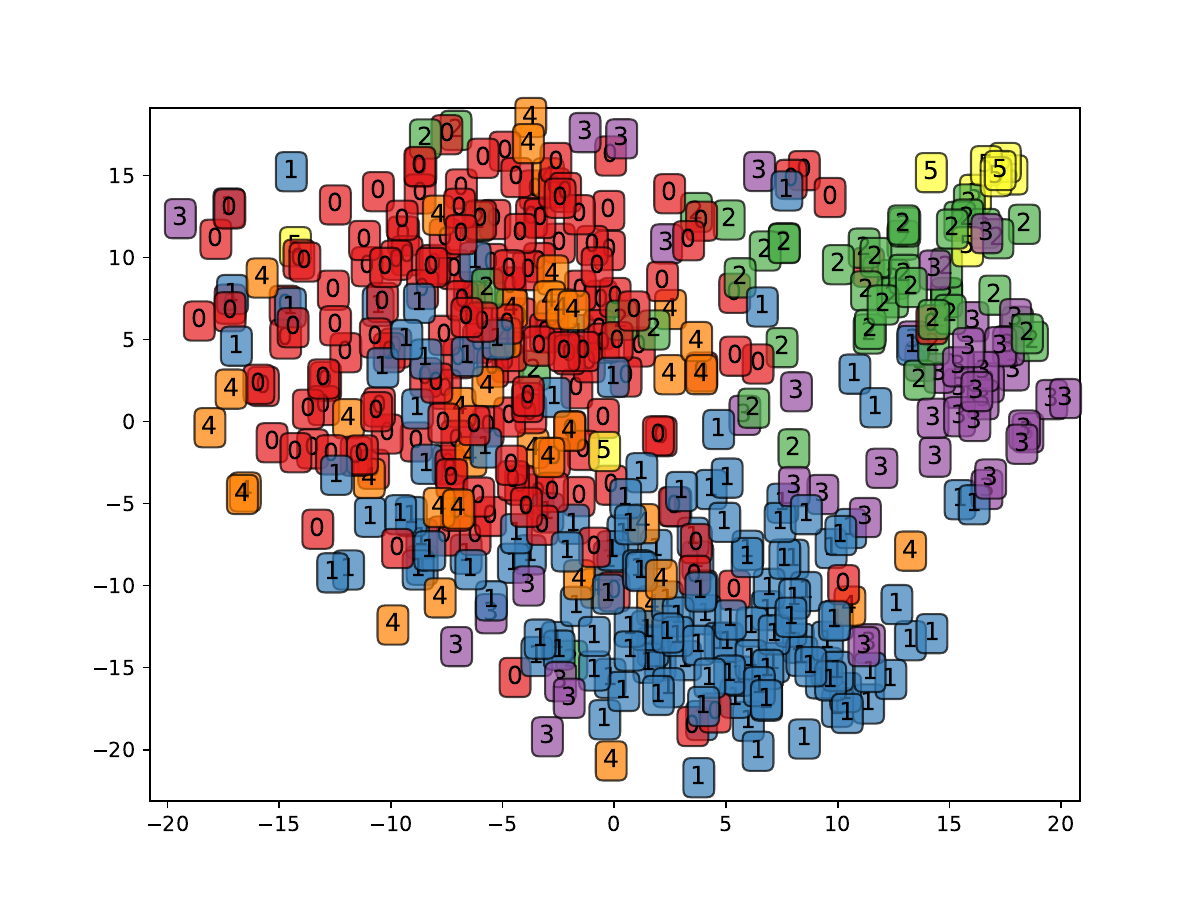}}
		\subfigure[JF-MMFA]{\includegraphics[width=4.2cm, height=5cm]{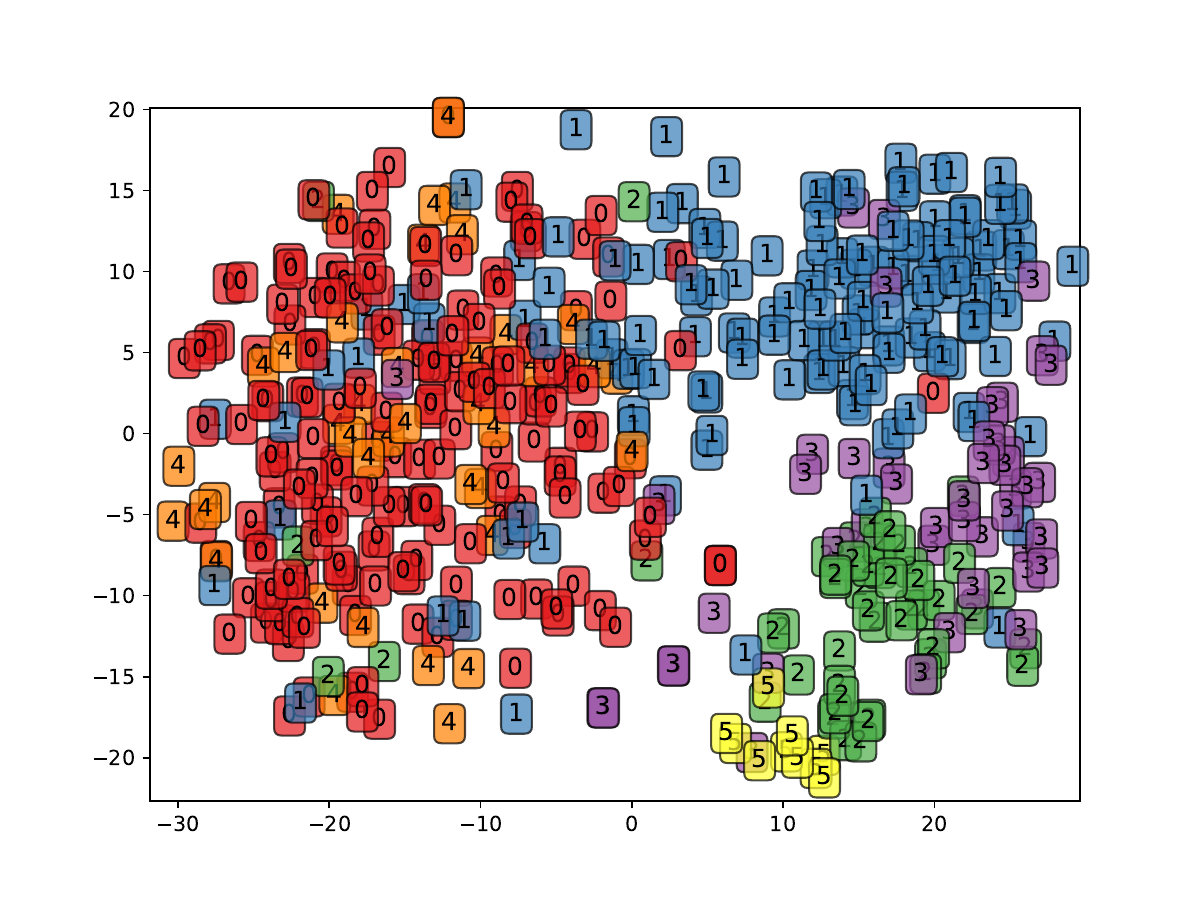}}
		\subfigure[JF-MA]{\includegraphics[width=4.2cm, height=5cm]{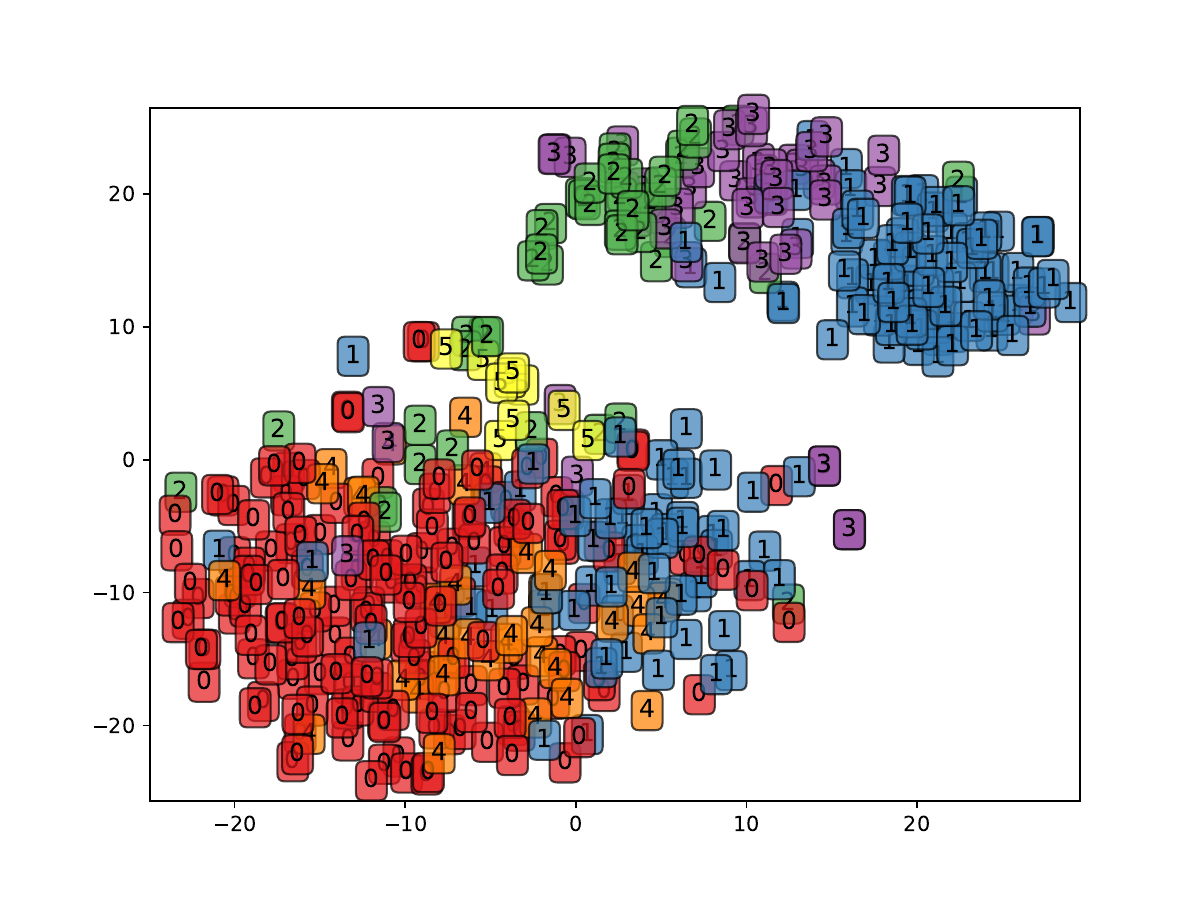}}
		
		\subfigure[JIF-MMFA(OFB)]{\includegraphics[width=4.2cm, height=5cm]{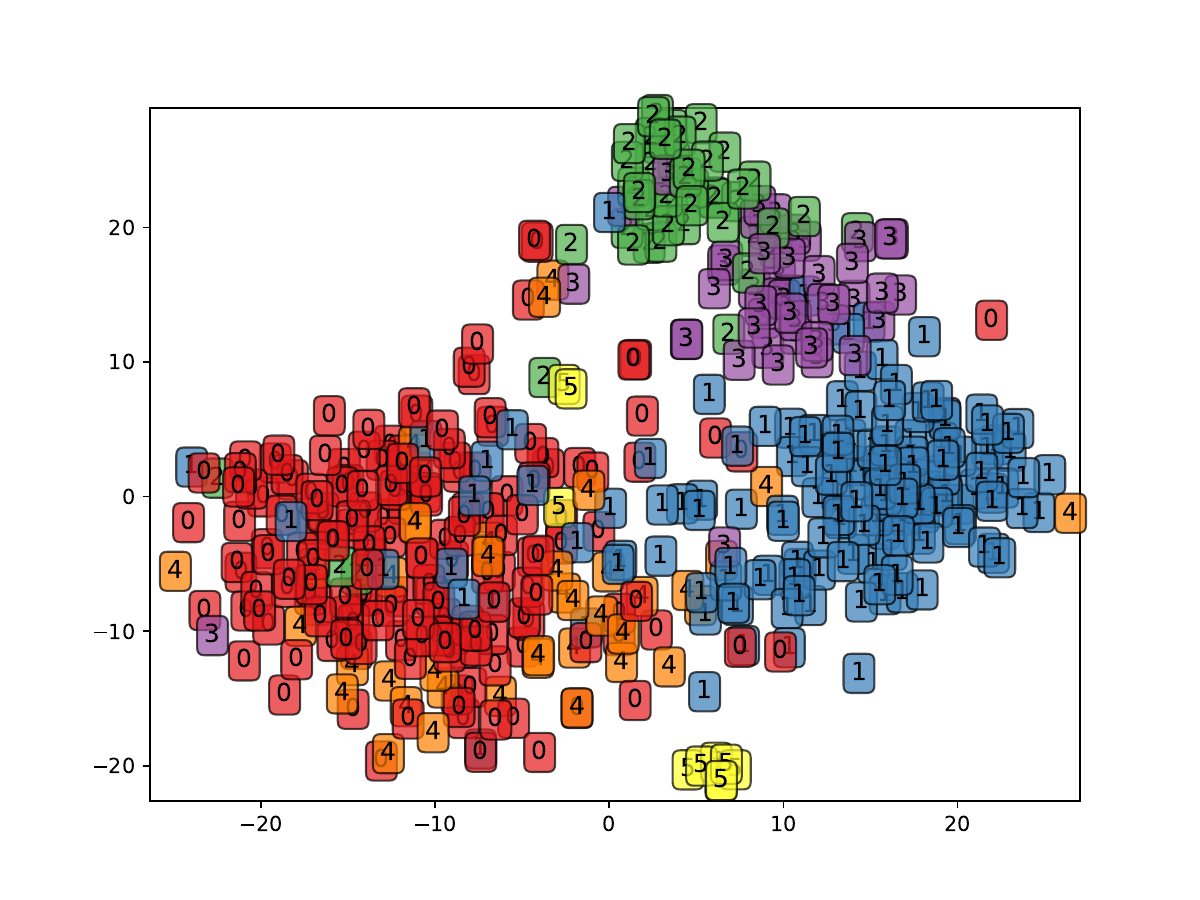}}		
		\subfigure[JIF-MMFA(All)]{\includegraphics[width=4.2cm, height=5cm]{./figures/new_fig.5/TSNE/JointLateFusion_FA_densenet_averaged.pdf}}

		\caption{\textcolor{black}{The T-SNE figures of the methods in Table~\ref{table1} considering  DenseNet-121 on the PAD-UFES-20 dataset. Here, 0-BCC: Basal Cell Carcinoma, 1-ACK: Actinic Keratosis, 2-NEV: Nevus, 3-SEK: Seborrheic Keratosis, 4-SCC: Squamous Cell Carcinoma and 5-MEL: Melanoma. See also sections 3.3.2 and 4.2.}} 
		
		\vspace{-0.2in}
		\label{fig5}
	\end{figure}

	\begin{table}[th]
		\centering
		\setlength{\abovecaptionskip}{0pt}
		\setlength{\belowcaptionskip}{10pt}		
			
		\caption{Performance comparison of different fusion structures (FS) with different fusion modules and CNN backbones on the PAD-UFES-20 dataset in terms of the BAC value. 
			The highest are highlighted in bold for each row. $FS$: Fusion structure; $Cat$: Concatenation; $MB$: Metablock; $MN$: MetaNet; $MA$: Mutual Attention, $MMFA$ Multi-Modal Fusion Attention. 
			JIF (OFB): the result only from the $P_{IM}$ of the JIF structure; 
			JIF (All): the result by averaging these three predictions $P_I$, $P_M$ and $P_{IM}$ of the JIF structure.
			(numbers in $\%$) }
		
		\scalebox{0.65}{
			\renewcommand\arraystretch{1.75}
			\setlength{\tabcolsep}{1mm}{

	\begin{tabular}{|c|ccccc|ccccc|}
		\hline
		\textit{\textbf{FS}}        & \multicolumn{5}{c|}{\textit{\textbf{JIF (OFB) }}}                                                                                                                                                               & \multicolumn{5}{c|}{\textit{\textbf{JIF (ALL)}}}                                                                                                                                                               \\ \hline
		\textit{\textbf{CNN}}       & \multicolumn{1}{c|}{\textit{\textbf{CAT}}} & \multicolumn{1}{c|}{\textit{\textbf{MB}}} & \multicolumn{1}{c|}{\textit{\textbf{MN}}} & \multicolumn{1}{c|}{\textit{\textbf{MA}}} & \textit{\textbf{MMFA}} & \multicolumn{1}{c|}{\textit{\textbf{CAT}}} & \multicolumn{1}{c|}{\textit{\textbf{MB}}} & \multicolumn{1}{c|}{\textit{\textbf{MN}}} & \multicolumn{1}{c|}{\textit{\textbf{MA}}} & \textit{\textbf{MMFA}} \\ \hline
		\textit{\textbf{densenet}}  & \multicolumn{1}{c|}{73.4$\pm$1.4}             & \multicolumn{1}{c|}{72.7$\pm$2.1}            & \multicolumn{1}{c|}{67.3$\pm$2.8}            & \multicolumn{1}{c|}{75.7$\pm$1.3}            & \textbf{78.0$\pm$2.0}     & \multicolumn{1}{c|}{74.0$\pm$1.2}             & \multicolumn{1}{c|}{74.8$\pm$1.3}            & \multicolumn{1}{c|}{69.7$\pm$2.0}            & \multicolumn{1}{c|}{76.3$\pm$1.5}            & 77.7$\pm$1.8              \\ \hline
		\textit{\textbf{mobilenet}} & \multicolumn{1}{c|}{74.7$\pm$2.5}             & \multicolumn{1}{c|}{71.7$\pm$3.8}            & \multicolumn{1}{c|}{68.5$\pm$1.9}            & \multicolumn{1}{c|}{75.7$\pm$0.8}            & 74.7$\pm$1.4              & \multicolumn{1}{c|}{75.7$\pm$2.8}             & \multicolumn{1}{c|}{73.6$\pm$2.1}            & \multicolumn{1}{c|}{71.1$\pm$1.2}            & \multicolumn{1}{c|}{\textbf{76.9$\pm$1.2}}   & 75.6$\pm$0.7              \\ \hline
		\textit{\textbf{resnet}}    & \multicolumn{1}{c|}{73.2$\pm$1.4}             & \multicolumn{1}{c|}{71.3$\pm$2.9}            & \multicolumn{1}{c|}{69.7$\pm$1.8}            & \multicolumn{1}{c|}{75.1$\pm$1.9}            & 76.0$\pm$1.2              & \multicolumn{1}{c|}{73.9$\pm$1.6}             & \multicolumn{1}{c|}{73.4$\pm$0.7}            & \multicolumn{1}{c|}{71.5$\pm$2.2}            & \multicolumn{1}{c|}{75.4$\pm$1.8}            & \textbf{76.4$\pm$1.5}     \\ \hline
		\textit{\textbf{effnet}}    & \multicolumn{1}{c|}{76.6$\pm$1.9}             & \multicolumn{1}{c|}{69.8$\pm$2.3}            & \multicolumn{1}{c|}{65.7$\pm$0.4}            & \multicolumn{1}{c|}{75.0$\pm$1.8}            & 78.8$\pm$1.6              & \multicolumn{1}{c|}{77.3$\pm$1.5}             & \multicolumn{1}{c|}{72.7$\pm$1.4}            & \multicolumn{1}{c|}{71.1$\pm$2.3}            & \multicolumn{1}{c|}{77.0$\pm$1.3}            & \textbf{79.8$\pm$1.4}     \\ \hline
		\textit{\textbf{xception}}  & \multicolumn{1}{c|}{74.2$\pm$1.4}             & \multicolumn{1}{c|}{70.5$\pm$0.4}            & \multicolumn{1}{c|}{65.2$\pm$2.8}            & \multicolumn{1}{c|}{75.3$\pm$2.2}            & 75.9$\pm$1.4              & \multicolumn{1}{c|}{74.8$\pm$1.6}             & \multicolumn{1}{c|}{72.9$\pm$0.9}            & \multicolumn{1}{c|}{68.8$\pm$2.7}            & \multicolumn{1}{c|}{75.6$\pm$1.8}            & \textbf{76.3$\pm$1.2}     \\ \hline
		\textit{\textbf{Average}}   & \multicolumn{1}{c|}{74.4$\pm$2.2}             & \multicolumn{1}{c|}{71.2$\pm$2.7}            & \multicolumn{1}{c|}{67.3$\pm$2.7}            & \multicolumn{1}{c|}{75.4$\pm$1.7}            & 76.7$\pm$2.2              & \multicolumn{1}{c|}{75.1$\pm$2.2}             & \multicolumn{1}{c|}{73.5$\pm$1.5}            & \multicolumn{1}{c|}{70.4$\pm$2.4}            & \multicolumn{1}{c|}{76.3$\pm$1.7}            & \textbf{77.2$\pm$2.0}     \\ \hline
	\end{tabular}

		}}
		\label{table7}
	\end{table}
	
	\begin{table}[th]
		\centering
		\setlength{\abovecaptionskip}{0pt}
		\setlength{\belowcaptionskip}{10pt}		
			
		\caption{Performance comparison of different fusion structures (FS) with different fusion modules and CNN backbones on the SPC dataset in terms of the BAC value. 
			The highest BAC values are highlighted in bold for each row. $FS$: Fusion structure; $Cat$: Concatenation; $MB$: Metablock; $MN$: MetaNet; $MA$: Mutual Attention, $MMFA$ Multi-Modal Fusion Attention. 
			JIF (OFB): the result only from the $P_{IM}$ of the JIF structure; 
			JIF (All): the result by averaging these three predictions $P_I$, $P_M$ and $P_{IM}$ of the JIF structure.
			(numbers in $\%$) }
		
		\scalebox{0.65}{
			\renewcommand\arraystretch{1.75}
			\setlength{\tabcolsep}{1mm}{
				
	\begin{tabular}{|c|ccccc|ccccc|}
		\hline
		\textit{\textbf{FS}}        & \multicolumn{5}{c|}{\textit{\textbf{JIF (OFB)}}}                                                                                                                                                               & \multicolumn{5}{c|}{\textit{\textbf{JIF (ALL)}}}                                                                                                                                                               \\ \hline
		\textit{\textbf{CNN}}       & \multicolumn{1}{c|}{\textit{\textbf{CAT}}} & \multicolumn{1}{c|}{\textit{\textbf{MB}}} & \multicolumn{1}{c|}{\textit{\textbf{MN}}} & \multicolumn{1}{c|}{\textit{\textbf{MA}}} & \textit{\textbf{MMFA}} & \multicolumn{1}{c|}{\textit{\textbf{CAT}}} & \multicolumn{1}{c|}{\textit{\textbf{MB}}} & \multicolumn{1}{c|}{\textit{\textbf{MN}}} & \multicolumn{1}{c|}{\textit{\textbf{MA}}} & \textit{\textbf{MMFA}} \\ \hline
		\textit{\textbf{densenet}}  & \multicolumn{1}{c|}{64.4$\pm$4.1}             & \multicolumn{1}{c|}{69.8$\pm$2.9}            & \multicolumn{1}{c|}{56.5$\pm$4.4}            & \multicolumn{1}{c|}{69.4$\pm$4.3}            & 70.9$\pm$2.3              & \multicolumn{1}{c|}{66.1$\pm$3.9}             & \multicolumn{1}{c|}{73.0$\pm$2.1}            & \multicolumn{1}{c|}{60.9$\pm$4.3}            & \multicolumn{1}{c|}{70.8$\pm$3.5}            & \textbf{73.1$\pm$2.6}     \\ \hline
		\textit{\textbf{mobilenet}} & \multicolumn{1}{c|}{68.0$\pm$2.4}             & \multicolumn{1}{c|}{68.9$\pm$1.8}            & \multicolumn{1}{c|}{62.5$\pm$2.9}            & \multicolumn{1}{c|}{74.5$\pm$2.3}            & 72.1$\pm$4.9              & \multicolumn{1}{c|}{68.9$\pm$1.9}             & \multicolumn{1}{c|}{72.2$\pm$1.9}            & \multicolumn{1}{c|}{65.6$\pm$2.9}            & \multicolumn{1}{c|}{\textbf{75.0$\pm$2.8}}   & 73.1$\pm$3.9              \\ \hline
		\textit{\textbf{resnet}}    & \multicolumn{1}{c|}{61.1$\pm$1.8}             & \multicolumn{1}{c|}{68.7$\pm$2.5}            & \multicolumn{1}{c|}{53.9$\pm$3.2}            & \multicolumn{1}{c|}{68.3$\pm$3.9}            & 70.0$\pm$2.7              & \multicolumn{1}{c|}{63.6$\pm$2.4}             & \multicolumn{1}{c|}{\textbf{71.7$\pm$3.0}}   & \multicolumn{1}{c|}{60.4$\pm$2.7}            & \multicolumn{1}{c|}{69.1$\pm$3.4}            & 70.4$\pm$2.6              \\ \hline
		\textit{\textbf{effnet}}    & \multicolumn{1}{c|}{74.7$\pm$1.3}             & \multicolumn{1}{c|}{69.7$\pm$2.1}            & \multicolumn{1}{c|}{53.4$\pm$4.0}            & \multicolumn{1}{c|}{70.8$\pm$2.2}            & 71.2$\pm$2.0              & \multicolumn{1}{c|}{\textbf{75.1$\pm$1.4}}    & \multicolumn{1}{c|}{71.6$\pm$1.3}            & \multicolumn{1}{c|}{65.9$\pm$2.0}            & \multicolumn{1}{c|}{72.2$\pm$1.8}            & 74.0$\pm$1.1              \\ \hline
		\textit{\textbf{xception}}  & \multicolumn{1}{c|}{70.6$\pm$1.6}             & \multicolumn{1}{c|}{67.5$\pm$1.1}            & \multicolumn{1}{c|}{58.0$\pm$1.8}            & \multicolumn{1}{c|}{72.5$\pm$1.6}            & 70.6$\pm$2.0              & \multicolumn{1}{c|}{71.3$\pm$1.8}             & \multicolumn{1}{c|}{68.9$\pm$0.9}            & \multicolumn{1}{c|}{66.9$\pm$1.4}            & \multicolumn{1}{c|}{\textbf{73.0$\pm$1.9}}   & 71.5$\pm$2.7              \\ \hline
		\textit{\textbf{Average}}   & \multicolumn{1}{c|}{67.7$\pm$5.3}             & \multicolumn{1}{c|}{68.9$\pm$2.3}            & \multicolumn{1}{c|}{56.9$\pm$4.7}            & \multicolumn{1}{c|}{71.1$\pm$3.8}            & 70.9$\pm$3.1              & \multicolumn{1}{c|}{69.0$\pm$4.7}             & \multicolumn{1}{c|}{71.5$\pm$2.4}            & \multicolumn{1}{c|}{63.9$\pm$3.9}            & \multicolumn{1}{c|}{72.1$\pm$3.4}            & \textbf{72.4$\pm$3.0}     \\ \hline
	\end{tabular}

		}}
		\label{table8}
	\end{table}
	
	\begin{table}[th]
		\centering
		\setlength{\abovecaptionskip}{0pt}
		\setlength{\belowcaptionskip}{10pt}		
			
		\caption{Performance comparison of different fusion structures (FS) with different fusion modules and CNN backbones on the ISIC-2019 dataset in terms of the BAC value. 
			The highest are highlighted in bold for each row. $FS$: Fusion structure; $Cat$: Concatenation; $MB$: Metablock; $MN$: MetaNet; $MA$: Mutual Attention, $MMFA$ Multi-Modal Fusion Attention. 
			JIF (OFB): the result only from the $P_{IM}$ of the JIF structure; 
			JIF (All): the result by averaging these three predictions $P_I$, $P_M$ and $P_{IM}$ of the JIF structure.
			(numbers in $\%$) }
		
		\scalebox{0.65}{
			\renewcommand\arraystretch{1.75}
			\setlength{\tabcolsep}{1mm}{

	\begin{tabular}{|c|ccccc|ccccc|}
		\hline
		\textit{\textbf{FS}}        & \multicolumn{5}{c|}{\textit{\textbf{JIF (OFB)}}}                                                                                                                                                                             & \multicolumn{5}{c|}{\textit{\textbf{JIF (ALL)}}}                                                                                                                                                                             \\ \hline
		\textit{\textbf{CNN}}       & \multicolumn{1}{c|}{\textit{\textbf{CAT}}} & \multicolumn{1}{c|}{\textit{\textbf{MB}}} & \multicolumn{1}{c|}{\textit{\textbf{MN}}} & \multicolumn{1}{c|}{\textit{\textbf{MA}}} & \multicolumn{1}{c|}{\textit{\textbf{MMFA}}} & \multicolumn{1}{c|}{\textit{\textbf{CAT}}} & \multicolumn{1}{c|}{\textit{\textbf{MB}}} & \multicolumn{1}{c|}{\textit{\textbf{MN}}} & \multicolumn{1}{c|}{\textit{\textbf{MA}}} & \multicolumn{1}{c|}{\textit{\textbf{MMFA}}} \\ \hline
		\textit{\textbf{densenet}}  & \multicolumn{1}{l|}{82.4$\pm$0.6}             & \multicolumn{1}{l|}{82.8$\pm$1.1}            & \multicolumn{1}{l|}{81.3$\pm$1.7}            & \multicolumn{1}{l|}{84.3$\pm$0.3}            & \textbf{84.8$\pm$1.1}                          & \multicolumn{1}{l|}{82.4$\pm$0.7}             & \multicolumn{1}{l|}{82.4$\pm$1.0}            & \multicolumn{1}{l|}{81.6$\pm$1.7}            & \multicolumn{1}{l|}{84.5$\pm$0.4}            & 84.6$\pm$0.9                                   \\ \hline
		\textit{\textbf{mobilenet}} & \multicolumn{1}{l|}{81.8$\pm$0.7}             & \multicolumn{1}{l|}{81.7$\pm$1.5}            & \multicolumn{1}{l|}{82.0$\pm$0.3}            & \multicolumn{1}{l|}{85.6$\pm$0.3}            & 85.0$\pm$1.5                                   & \multicolumn{1}{l|}{81.9$\pm$0.8}             & \multicolumn{1}{l|}{81.4$\pm$1.2}            & \multicolumn{1}{l|}{82.0$\pm$0.6}            & \multicolumn{1}{l|}{\textbf{85.7$\pm$0.2}}   & 84.8$\pm$1.4                                   \\ \hline
		\textit{\textbf{resnet}}    & \multicolumn{1}{l|}{82.8$\pm$0.2}             & \multicolumn{1}{l|}{82.0$\pm$0.7}            & \multicolumn{1}{l|}{81.7$\pm$1.0}            & \multicolumn{1}{l|}{\textbf{84.3$\pm$0.7}}   & 83.7$\pm$0.5                                   & \multicolumn{1}{l|}{82.8$\pm$0.4}             & \multicolumn{1}{l|}{81.5$\pm$0.7}            & \multicolumn{1}{l|}{82.0$\pm$1.1}            & \multicolumn{1}{l|}{84.1$\pm$0.5}            & 83.7$\pm$0.3                                   \\ \hline
		\textit{\textbf{effnet}}    & \multicolumn{1}{l|}{79.0$\pm$1.6}             & \multicolumn{1}{l|}{79.6$\pm$1.0}            & \multicolumn{1}{l|}{78.6$\pm$1.0}            & \multicolumn{1}{l|}{82.5$\pm$0.5}            & \textbf{82.6$\pm$0.6}                          & \multicolumn{1}{l|}{79.0$\pm$1.7}             & \multicolumn{1}{l|}{79.9$\pm$0.7}            & \multicolumn{1}{l|}{79.1$\pm$1.0}            & \multicolumn{1}{l|}{\textbf{82.7$\pm$0.9}}   & 82.5$\pm$0.7                                   \\ \hline
		\textit{\textbf{xception}}  & \multicolumn{1}{l|}{78.6$\pm$1.3}             & \multicolumn{1}{l|}{79.0$\pm$0.9}            & \multicolumn{1}{l|}{78.0$\pm$0.7}            & \multicolumn{1}{l|}{82.3$\pm$0.3}            & 82.5$\pm$0.3                                   & \multicolumn{1}{l|}{78.7$\pm$1.3}             & \multicolumn{1}{l|}{79.0$\pm$0.8}            & \multicolumn{1}{l|}{78.5$\pm$0.5}            & \multicolumn{1}{l|}{82.2$\pm$0.4}            & \textbf{82.7$\pm$0.3}                          \\ \hline
		\textit{\textbf{Average}}   & \multicolumn{1}{l|}{80.9$\pm$2.0}             & \multicolumn{1}{l|}{81.0$\pm$1.8}            & \multicolumn{1}{l|}{80.3$\pm$2.0}            & \multicolumn{1}{l|}{\textbf{83.8$\pm$1.3}}   & \textbf{83.8$\pm$1.4}                          & \multicolumn{1}{l|}{80.9$\pm$2.1}             & \multicolumn{1}{l|}{80.9$\pm$1.5}            & \multicolumn{1}{l|}{80.6$\pm$1.9}            & \multicolumn{1}{l|}{\textbf{83.8$\pm$1.3}}   & 83.7$\pm$1.3                                   \\ \hline
	\end{tabular}

		}}
		\label{table9}
	\end{table}	
	
	\subsubsection{Effectiveness of using patient's metadata}
	
	The experiments in this part show the effectiveness of using the patient metadata in addition to the image data.
	As shown in Table~\ref{table1}, Table~\ref{table2}, and Table~\ref{table3}, the models that use metadata all obtain higher BAC values than the models that do not use metadata on all three datasets. 
	Particularly, when the proposed JIF-MMFA (All) method is applied, a significant improvement for all five CNN backbones and multiple datasets is obtained. 
	Compared with the models that only use images, our JIF-MMFA (All) remarkably improves the average value from $67.0\%\pm2.3\%$ to $77.2\%\pm2.0\%$ on the PAD-UFES-20 dataset, $55.4\%\pm3.8\%$ to $72.4\%\pm3.0\%$ on the SPC dataset, and $80.4\%\pm1.5\%$ to $83.7\%\pm1.3\%$ on the ISIC-2019 dataset.
	However, JF-MN only get a slight increase of $0.8\%$ on PAD-UFES-20 dataset.
	These results demonstrate that compared to the model only using image, fusing patient metadata and images can boost the performance, while the performance improvement depends on the fusion methods, and our proposed JIF-MMFA makes the most improvement.

	\subsubsection{Performance comparison between our JIF-MMFA method and other fusion methods}
		
	In this part, we investigate the performance improvement brought by the JIF structure and the MMFA module in our JIF-MMFA method, we report our method's ablation study, i.e. the performance comparison of JF-MMFA, JIF-MMFA (OFB) and JIF-MMFA (All). 
	Then, to demonstrate the advantage of JIF-MMFA, we compare the proposed JIF-MMFA approach with the other fusion methods (JF-CAT, JF-MB, and JF-MN) on the three datasets.
	
	First, compared with the JF structure, our JIF structure preserves the specific properties of each modality to learn a better joint feature presentation, and integrates the multi-modal information at the decision level.
	To figure out the performance improvement brought by these two factors, we compare the results obtained from only-fusion-branch (OFB) $P_{IM}$ of the JIF structure (JIF-MMFA (OFB)), and that from the average of all three predictions of JIF structure (JIF-MMFA (All)): $P_{IM}$, $P_{I}$ and $P_M$ (see Fig.~\ref{fig1}(b)), see Table~\ref{table1}, Table~\ref{table2}, and Table~\ref{table3}.
	Compared with JF-MMFA, JIF-MMFA (OFB) increases the averaged BAC value from $74.9\%\pm2.3\%$ to $76.7\%\pm2.2\%$ on the PAD-UFES-20 dataset, from $69.6\%\pm3.2\%$ to $71.0\%\pm3.1\%$ on the SPC dataset, and from $78.9\%\pm5.9\%$ to $83.7\%\pm1.3\%$ on the ISIC-2019 dataset.
	JIF-MMFA (All) gets a slightly higher value in averaged BAC than JIF-MMFA (OFB) on these three datasets.
	These results illustrate that the improvement of JF-MMFA and JIF-MMFA is mainly from the preservation of modal-specific features that can learn a better joint feature representation, and not much influenced by the decision-level fusion of multi-modality data.
	
	Second, we compare JIF-MMFA with the other fusion methods: Joint Fusion structure with Concatenation (JF-CAT), Metablock (JF-MB), Metanet (JF-MN), and Mutual Attention (JF-MA), \citep{li2020fusing,pacheco2020impact,pacheco2021attention,cai2022multimodal}.
	The proposed JIF-MMFA (All) method outperforms all the other methods on all the datasets according to the average BAC value. 
	Compared to previous methods, our JIF-MMFA (All) achieves a significant improvement on both PAD-UFES-20 dataset and SPC dataset, i.e., an increase of $2.7\%$ in averaged BAC value compared with the second-best method (JF-MA) on the PAD-UFES-20 dataset (see Table~\ref{table1}), and an increase of $3.5\%$ on the SPC dataset (see Table~\ref{table2}), However, on the ISIC-2019 dataset, JIF-MMFA (OFB) ($83.7\%\pm1.3\%$) achieves an increase of $1.9\%$ in averaged BAC value compared with the second-best method (JF-CAT $81.8\%\pm1.3\%$), which demonstrate the advantage of our method.
	
	The Friedman test and Wilcoxon test is also performed for statistical analysis, using $p = 0.05$.
	The $p$ value obtained by the Friedman test is about $1.99 \times 10^{-24}$, $8.71 \times 10^{-23}$, and $1.06 \times 10^{-12}$ on the PAD-UFES-20 dataset, the SPC dataset and the ISIC-2019 dataset, respectively. 
	Thus, we continue to conduct the Wilcoxon test (two-sided), and show these results in Table~\ref{table4}, Table~\ref{table5}, and Table~\ref{table6}).
	From these tables, we can that see on the PAD-UFES-20 and SPC datasets, the model-pairs of our JIF-MMFA (ALL) with previous methods (JF-CAT, JF-MB, JF-MN, JF-MA), all the return values are greater than 0.05, which illustrate the JIF-MMFA (ALL) generally performs better that these methods.	
	The confusion matrix and the T-SNE plot of the different fusion methods are depicted in Fig.~\ref{fig4} and Fig.~\ref{fig5}. 
	As there are 15 confusion matrices and T-SNE figures, we decided to only display the result of DenseNet-121 on the PAD-UFES-20 dataset, because it is a lightweight and very common CNN backbone in deep learning and it presents a fair performance in our experiments.
	It can be seen that the relationship between Fig.~\ref{fig4} and Fig.~\ref{fig5} that the higher misclassification rate between two types skin diseases in Fig.\ref{fig4}, the more close distance between the corresponding two types in Fig.~\ref{fig5}. For example, in the case of our JIF-MMFA (All) model, 38$\%$ SCC cases are predicted to BCC in Fig.\ref{fig4} (g), so we can the cluster between 0 (BCC) and 4 (SCC) is hard separate in Fig.\ref{fig5}, however, there are no BCC cases were predicted to NEV and SEK in Fig.\ref{fig4} (g), as the cluster between 0 (BCC) and 2 (NEV), 3 (SEK) are almost separated. It is because the features we use to conduct TSNE is the final feature vector of the fusion method, which is directly used for prediction.

	\subsubsection{Effectiveness of Joint-Individual Fusion (JIF) structure}

	To further evaluate the effectiveness of our JIF structure, we compare the JIF structure with the JF structure using four different fusion modules (see Table~\ref{table7}, Table~\ref{table8}, and Table~\ref{table9}) . 
	On the PAD-UFES-20 dataset (see Table~\ref{table7}), it can be seen that JIF (All) structure improves the performance of all the four fusion modules in terms of the average BAC value, and JIF (OFB) improves in 3 out of 4 fusion modules except Metanet, compared with the JF structure.
	Like for the PAD-UFES-20 dataset, both JIF (All) and JIF (OFB) improve all four fusion modules on the SPC dataset, see Table~\ref{table8}.
	On the ISIC-2019 dataset (see Table~\ref{table9}), JIF (All) and JIF (OFB) improve the performance of the MA fusion module from $78.7\%\pm7.9\%$ to $83.8\%\pm1.3\%$ and to $83.8\%\pm1.3\%$, respectively, and MMFA fusion module from $78.9\%\pm5.5\%$ to $83.6\%\pm1.3\%$ and to $83.8\%\pm1.2\%$, respectively, while degenerating the performance of MN fusion module from $81.7\%\pm2.1\%$ to $80.3\%\pm2.0$ and to $80.6\%\pm1.9\%$, respectively, and the CAT fusion module from $81.8\%\pm1.7\%$ to $80.9\%\pm2.0$ and $80.9\%\pm2.1\%$, respectively.
	
	To conclude, compared with the JF structure, all the fusion modules are improved by the JIF (All) structure on the PAD-UFES-20 and SPC datasets (the datasets with more types of metadata) in terms of averaged BAC value, but are affected by the JIF (All) structure on ISIC-2019 dataset (the dataset with less types of metadata), except for MA and MMFA module. 
	It proves the generalization ability of the JIF structure to all fusion modules on the datasets with more metadata.
	Also, these results suggest that our JIF structure may be less effective for other fusion modules - the fusion modules that only use metadata to enhance image feature or conduct simple transformations - on the dataset with few metadata.

	\subsubsection{Effectiveness of Multi-Modal Fusion Attention (MMFA) Module}
	
	To show the effectiveness of the proposed MMFA module, we compare it with the other three fusion modules (CAT, MB, MN and MA) combined with different fusion structures and on the different datasets.
	As shown in Table~\ref{table1} and Table~\ref{table2}, Table~\ref{table7} and Table~\ref{table8}, our MMFA module consistently obtains the highest average BAC value when embedded in all the three fusion structures: JF, JIF (OFB) and JIF (All), which are $74.9\%\pm2.3\%$, $76.7\%\pm2.2\%$, and $77.2\%\pm2.0\%$, respectively, on the PAD-UFES-20 dataset, and $69.6\%\pm3.2\%$, $71.0\%\pm3.1\%$, and $72.4\%\pm3.0\%$, respectively, on the SPC dataset.
	As displayed in Table~\ref{table3}, compared with the other fusion modules, MA achieves the lowest BAC value of $78.7\%\pm7.9\%$ and MMFA achieves the second-lowest BAC value of $78.9\%\pm5.8\%$ when combined with the JF structure on the ISIC-2019 dataset.
	However, as shown in Table~\ref{table9}, these two modules obtain the top-2 ranking BAC value with the JIF (OFB) ($83.8\%\pm1.3\%$ and $83.8\%\pm1.4\%$) and JIF (All) ($83.8\%\pm1.3\%$ and $83.7\%\pm1.3\%$) structures. 

	In total, these results illustrate that in Table~\ref{table7}, Table~\ref{table8}, and Table~\ref{table9} it is shown that the proposed MMFA has a higher averaged BAC value than the other fusion modules based on the JF and JIF (All) structures.

	\section{Discussion}
	
	\subsection{Effectiveness of using patient's metadata}
	From the results of Table~\ref{table1}, Table~\ref{table2}, and Table~\ref{table3}, it can be seen that compared with the model not using metadata, JIF-MMFA (All) achieves much more improvements on the PAD-UFES-20 dataset and the SPC dataset, while it only achieves an insignificant improvement on the ISIC-2019 dataset. 
	We believe that this is because the 21 and 14 metadata features of the PAD-UFES-20 and SPC datasets are more valuable, while the ISIC-dataset only has very limited patient metadata, such as age, location, and gender. 
	
	\subsection{Performance comparison between our JIF-MMFA method and other fusion methods}
	Further, JIF-MMFA (All) achieves the highest BAC value in 12 out of 15 CNN scenarios, except for Xception on the SPC dataset and Resnet-50 and Efficientnet-B3 on the ISIC-2019 dataset. 
	However, in the above-mentioned three scenarios, JIF-MMFA (All) also achieves comparable performance with the best fusion methods, proving our method's generalization ability for CNNs. 
	For instance, in the case of Efficientnet-B3 on the SPC dataset, there are only subtle gaps between the BAC values that are obtained by the best-performing method JF-CAT and our JIF-MMFA (All).
	
	The statistical results in Table~\ref{table4}, Table~\ref{table5}, and Table~\ref{table6}  show that JIF-MMFA (All) with other fusion methods (not including JF-MMFA and JIF-MMFA (OFB)) return $p<0.05$, 
	which illustrate that JIF-MMFA generally performs better than other fusion methods.
	
	Next, the confusion matrices displayed in Fig.~\ref{fig4} present an interesting result. 
	Generally, the metadata assists the CNN model to increase the diagnostic rate of all the skin diseases.
	However, the mis-classification rate between BCC and SCC is still considerable. 
	This is because these two lesions have not only similar visual features, but also many similar values in the metadata. 
	In fact, classifying SCC and BCC is a challenging task even for experienced dermatologists with the use of dermoscopy. 
	Nevertheless, this confusion is not a big problem, as both are types of skin cancer and require biopsy for further evaluation. 
	It is a real problem to confuse them with ACK, which is a minor skin disease that is treated without a surgical process \citep{pacheco2021attention}.  
	What is more, it is worth noticing that the metadata helps distinguishing NEV from MEL, which is quite helpful for the expert's diagnosis, since NEV is benign, circumscribed malformations of the skin, while MEL is one of the most malignant cancers.
	For the classification of BCC, SCC, and ACK, JIF-MMFA (All) and JF-MB achieve better performance (see Fig.~\ref{fig4}).
	A similar phenomenon is also observed in the T-SNE figures (Fig.~\ref{fig5}) that JIF-MMFA (All) and JF-MB improve the clustering of samples between BCC, SCC, and ACK. 
	But it is still hard to differentiate the lesions in the sub-clusters. 
	It reflects the problem of inter-class similarity and intra-class variation for skin lesion classification.
	For MEL, our JIF-MMFA (All) method achieves the best performance according to the averaged BAC value.
	
	Also, it can be observed from the results for each CNN backbone using JIF-MMFA (All) that Efficient-B3 performs better than other backbones, and gets the most improvements than other CNN models compared the it only uses mage data on PAD-UFES-20 and SPC datasets, which somehow proves that the Efficient-B3 model is suitable as image model $I_M$ for multi-modal skin diseases classification.
	
	Finally, JIF-MMFA increases the parameters of the models not using metadata when it is applied to the CNN backbone, but the increase is not significant. 
	We follow the paper of \cite{pacheco2021attention} and only consider the experiments on the PAD-UFES-20 dataset, in which the number of model parameters of Densenet-121, Mobilenet-v2, Resnet-50, Efficientnet-B3, and Xception are increased by 0.08, 0.22, 0.04, 0.05, 0.08 and 0.05. 
	It seems that Mobilenet-v2 is the most impacted model with an increase of 0.22. 
	However, JIF-MMFA only increases the Mobilenet-v2's parameters from $3.6\times10^{6}$ to $4.4\times10^{6}$, which is insignificant in terms of  training time.

	\subsection{Effectiveness of the Multi-Modal Fusion Attention (MMFA) Module}
	Some interesting phenomenon about MMFA in Table~\ref{table9} shows that that the MMFA module achieves the worst performance   
	when combined with JF, but the best performance when combined with JIF on the ISIC-2019 dataset. 
	These results and the characteristic of MMFA (the only fusion module that mutually enhances image and metadata features) suggest that the CNN with MMFA module cannot conduct the mutual attention mechanism on image and metadata features well when combined with JF structure on the dataset with few metadata (ISIC-2019 dataset). 
	Further considering the results of the JF and JIF structures in Table~\ref{table9}, we believe that  
	this problem of the JF structure can be handled by the JIF structure that well preserves the modal-specific feature.

	\section{Conclusion}
	\textbf{In this paper, we propose the Joint-Individual Fusion (JIF) structure with the Multi-Modal Fusion Attention (MMFA) module for skin cancer classification.
		Firstly, the proposed MMFA module simultaneously and mutually enhances the image and metadata features by efficiently utilizing mutli-head self-attention mechanism, and thus achieves the better performance than other attention modules.
		Secondly, compared with other methods that ignores the exploration on fusion structure, we conduct a comprehensive exploration of different fusion structure.
		Furthermore, we proposed a Joint-Indivual Fusion structure can learn better shared features by preserving modal-specific
		features, and thus boost the classification performance of all the fusion modules in the most of scenarios.
		The experimental results on three public datasets proves our proposed JIF-MMFA achieves the highest averaged BAC value on all three datasets and the effectiveness of JIF and MMFA respectively.
		What is more, the Friedman test and Wilcoxon test indicate that our method is statistically better on all the datasets. 
		The experimental results in ISIC-2019 dataset also show that compard with JF structure, our JIF structure can not improve the performance of non-mutual attention fusion modules (CAT, MB and MN) in the dataset with few metadata,
		Therefore,  our future work will focus on the research of adptive fusion structure that has a strong generalization ability on in different situations}

	\section*{Acknowledgement}
	The authors appreciate the creator of seven-point checklist dataset for the release and organization of this dataset.
	This work was partially supported by the China Scholarship Council and the German Federal Ministry of Health (2520DAT920).

	\bibliographystyle{elsarticle-harv} 
	\bibliography{myreference_11_08}
\end{document}